\documentclass[manuscript,acmsmall,natbib=false]{acmart}
\AtBeginDocument{%
  \providecommand\BibTeX{{%
    \normalfont B\kern-0.5em{\scshape i\kern-0.25em b}\kern-0.8em\TeX}}}

\setcopyright{rightsretained}
\acmJournal{TAAS}
\acmYear{2024}
\acmVolume{1}
\acmNumber{1}
\acmArticle{1}
\acmMonth{1}
\acmDOI{10.1145/3687130}

\RequirePackage[
  datamodel=acmdatamodel,
  style=acmnumeric,
  ]{biblatex}
  
\usepackage{soul}

\usepackage{subcaption}
\usepackage{tabularx}
\usepackage{amsmath}
\usepackage{arydshln}

\newcommand{\Hq}{\mathbf{H}}

\usepackage{dsfont}

\newcommand{\ind}[1]{\mathds{1}\left\{#1\right\}}

\newcommand\norm[1]{\left\lVert#1\right\rVert}
\newcommand\V[1]{\mathrm{\textbf{var}}\mleft[#1\mright]}
\newcommand{\BR}{\mathbf{R}}
\newcommand{\BQ}{\mathbf{Q}}
\newcommand{\BI}{\mathbf{I}}
\newcommand{\BO}{\mathbf{0}}
\newcommand{\SQ}{S_\BQ}
\newcommand{\SR}{S_\BR}

\usepackage{mleftright}
\makeatletter
\newcommand*{\eval}{%
	\def\E@sub{}%
	\def\E@sup{}%
	\E@scripts
}
\newcommand*{\E@scripts}{%
	\@ifnextchar_\E@subscript{%
		\@ifnextchar^\E@supscript\E@finish
	}%
}
\def\E@subscript_#1{%
	\ifx\E@sub\@empty
	\def\E@sub{#1}%
	\else
	\errmessage{E already has a subscript}%
	\fi
	\E@scripts
}
\def\E@supscript^#1{%
	\ifx\E@sup\@empty
	\def\E@sup{#1}%
	\else
	\errmessage{E already has a superscript}%
	\fi
	\E@scripts
}
\newcommand*{\E@finish}[1]{%
	\mathbb{E}%
	\ifx\E@sub\@empty\else _{\E@sub}\fi
	\ifx\E@sup\@empty\else ^{\E@sup}\fi
	\mleft[#1\mright]%
}
\makeatother

\usepackage{xcolor}
\usepackage{forest}
\usetikzlibrary{positioning,arrows.meta}
\forestset{
  declare keylist={extra forest edges}{},
  declare boolean={draw extra edges}{0},
  align middle child/.style={
    before typesetting nodes={
      if={
        > Ow+P {n children}{isodd(##1)}
      }{
        calign child/.process={
          Ow+n {n children}{(##1+1)/2}
        },
        calign=child edge,
      }{},
    },
  },
  align middle children/.style={
    for tree={align middle child},
  },
  join from/.style={
    if draw extra edges={}{draw extra forest edges},
    before typesetting nodes={temptoksa/.option=#1.name, extra forest edges/.register=temptoksa},
  },
  join to/.style={
    before typesetting nodes={temptoksa/.option=name, if={>O{#1.draw extra edges}}{}{#1.draw extra forest edges}, #1.extra forest edges/.register=temptoksa},
  },
  draw extra forest edges/.style={
    draw extra edges,
    before drawing tree={
     tikz+/.process={OOw2{edge}{extra forest edges}{\foreach \i in {##2} \path [##1] ({\i}.parent anchor) -- (.child anchor);}},
    },
  },
}

\begin{document}

\title{Optimizing Delegation in Collaborative Human-AI Hybrid Teams}

\author{Andrew Fuchs}
\email{andrew.fuchs@phd.unipi.it}
\orcid{0000-0001-7191-8781}
\affiliation{%
  \institution{Department of Computer Science, Universit\`{a} di Pisa}
  \city{Pisa}
  \country{Italy}
  \postcode{56124}
}
\affiliation{%
  \institution{Institute for Informatics and Telematics (IIT), National Research Council (CNR)}
  \city{Pisa}
  \country{Italy}
  \postcode{56124}
}

\author{Andrea Passarella}
\orcid{0000-0002-1694-612X}
\affiliation{%
  \institution{IIT, CNR}
  \city{Pisa}
  \country{Italy}
  \postcode{56124}
}

\author{Marco Conti}
\orcid{0000-0003-4097-4064}
\affiliation{%
  \institution{IIT, CNR}
  \city{Pisa}
  \country{Italy}
  \postcode{56124}
}

\renewcommand{\shortauthors}{Fuchs, et al.}

\begin{abstract}
When humans and autonomous systems operate together as what we refer to as a hybrid team, we of course wish to ensure the team operates successfully and effectively. We refer to team members as agents. In our proposed framework, we address the case of hybrid teams in which, at any time, only one team member (the control agent) is authorized to act as control for the team. To determine the best selection of a control agent, we propose the addition of an AI manager (via Reinforcement Learning) which learns as an outside observer of the team. The manager learns a model of behavior linking observations of agent performance and the environment/world the team is operating in, and from these observations makes the most desirable selection of a control agent. From our review of current state of the art, we present a novel manager model for oversight of hybrid teams by our support for diverse agents and decision-maker operation across multiple time steps and decisions. In our model, we restrict the manager's task by introducing a set of constraints. The manager constraints indicate acceptable team operation, so a violation occurs if the team enters a condition which is unacceptable and requires manager intervention. To ensure minimal added complexity or potential inefficiency for the team, the manager should attempt to minimize the number of times the team reaches a constraint violation and requires subsequent manager intervention. Therefore, our manager is optimizing its selection of authorized agents to boost overall team performance while minimizing the frequency of manager intervention. We demonstrate our manager's performance in a simulated driving scenario representing the case of a hybrid team of agents composed of a human driver and autonomous driving system. We perform experiments for our driving scenario with interfering vehicles, indicating the need for collision avoidance and proper speed control. Our results indicate a positive impact of our manager, with some cases resulting in increased team performance up to $\approx 187\%$ that of the best solo agent performance.
\end{abstract}

\ccsdesc[500]{Human-centered computing~Ubiquitous and mobile computing}
\ccsdesc[500]{Computing methodologies~Reinforcement learning}
\ccsdesc[500]{Computing methodologies~Cooperation and coordination}

\keywords{hybrid team; reinforcement learning; collaboration; delegation}

\received{8 February 2024}
\received[revised]{3 June 2024}
\received[accepted]{1 August 2024}

\maketitle

\begin{tikzpicture}[remember picture,overlay]
\node[anchor=north,yshift=-20pt] at (current page.north) {\fbox{\parbox{\dimexpr\textwidth-\fboxsep-\fboxrule\relax}{
  \footnotesize{
    This work has been accepted by the ACM Transactions on Autonomous and Adaptive Systems for publication. Copyright may be transferred without notice, after which this version may no longer be accessible.
  }
}}};
\end{tikzpicture}
\vspace{-15pt}

\section{Introduction}

As we continue to see an increase in performance and adoption of automated or autonomous systems (via artificial intelligence), we anticipate the need to consider cases of teams of humans and autonomous systems which cooperate to implement a given task. The humans and autonomous systems, both of which we call ``agents'', comprise what we refer to as a hybrid team. The use of hybrid teams will depend on the scenario, which impacts the way the team functions. With autonomous systems performing at sufficient levels for operation with no, or minimal, human oversight, tasks can be offloaded to autonomous agents. The offloading of tasks enables either concurrent action of multiple human and autonomous agents or sequential decisions made one at a time \cite{carroll2019utility,chen2018planning,10.1007/978-3-030-62056-1_44,westby2023collective,wu2021too}. In either case, the performance of an agent with respect to a task will inform the decision to delegate either a task or the decision authority to that agent. These methods can also enable the use of autonomous systems as a decision-making aid for the human user. This can be accomplished by reducing the size of selections for a user, providing the autonomous system's suggestion, and more. With a team operating in a manner involving task sharing (concurrent or sequential), careful consideration is required to account for potential agent errors. Both humans and autonomous systems have shown they can perform at high levels, but there is also evidence that these agents are capable of errors \cite{agudo2024impact,cabrera2021discovering,haegler2010no,mahmood2022owning,reason1990human,russell2017human}. The scope and severity of the errors can serve as a measure for agent suitability. Depending on the acceptable level of risk with respect to the agents, the choice to delegate authority/tasks can be made according to performance, expected error likelihoods, etc. Similarly, autonomous systems can be designed with the option to abstain from making a decision when they determine a human/expert is more suitable \cite{afanador2019adversarial,chen2023learning}.

\begin{figure}[ht]
\vspace{-1mm}
    \centering
    \includegraphics[width=0.7\textwidth]{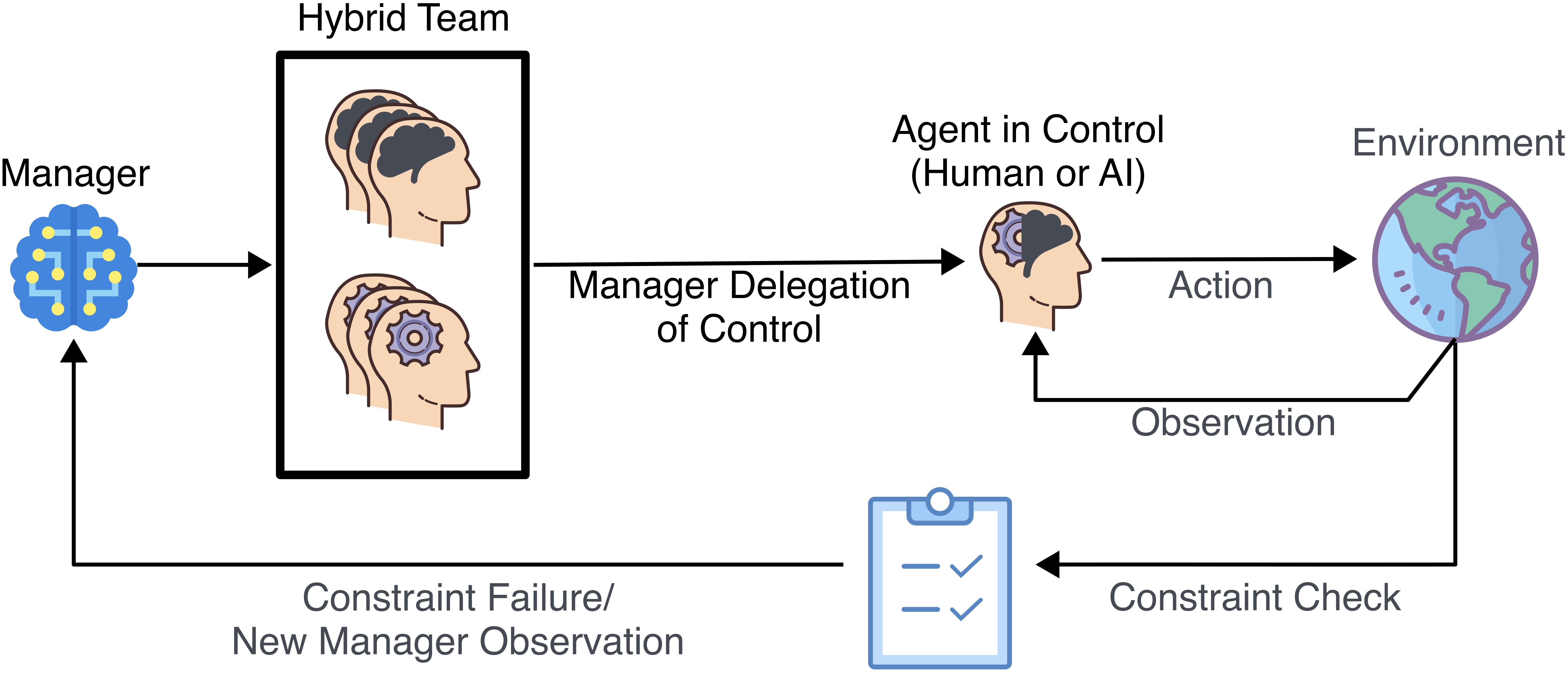}
    \caption{Collaborative hybrid team dynamic with single agent control. Controlling agent remains in control until constraint violation, which cues the manager to make new a delegation of authority.}
    \label{fig:managed_team_diagram}
\vspace{-2mm}
\end{figure}

In the previous examples of teaming or assistance, we see several common scenarios. First, teams can operate by splitting the tasks and concurrently accomplishing them. Second, humans and autonomous systems can be paired to share insights and make joint decisions. Last, a single decision for a team can be assigned to a single agent at each step of the process. While these represent relevant scenarios, we make a distinction to represent an alternate case. For our framework, we propose considering a slightly different form of single-agent delegation. In this case, the hybrid team consists of two or more human/autonomous agents under guidance of a manager, and the authority to make a decision is delegated by the manager to a single agent. To distinguish from the previously noted approaches, our delegation of authority is extended to allow agents to take actions until a new delegation is made. More specifically, the delegated agent will maintain authorization until the manager must make a new delegation. The manager is tasked specifically with identifying optimal delegations (see Figure~\ref{fig:managed_team_diagram}) when a constraint failure occurs, at which point the manager makes a new observation and makes a new delegation decision. By optimal, we will refer to aspects specific to team performance and how likely a delegated agent is to cause the need for manager intervention. A manager will make a delegation by intervening if the delegated agent transitions the team to a context which is undesirable from the manager's perspective. We will refer to these points at which the context is undesirable as critical points.

The single-agent operation for a multi-agent hybrid team allows us to support cases where there are multiple, possibly erroneous, decision-makers available and no concurrent behavior. By erroneous, we mean agents could have models of behavior that elicit outcomes which are undesirable in some cases. For example, in driving, our hybrid team would consist of a human driver and autonomous driving system. At any time, the world context/conditions (e.g., traffic conditions, time/light, weather, etc.) will all impact which agent is more suitable for delegation of control. For instance, there might be tasks such as parallel parking which can be better handled by an autonomous system than by most drivers. In this case, unless otherwise indicated by driver performance, it would make the most sense to delegate authority to the autonomous system for parking. However, adverse conditions could reduce autonomous system performance, and the human should be in control. With adverse conditions, we could see a human or autonomous driver make an error leading to a collision or other driving incident. Generalizing from the driving scenario, the common aspect is the manager selecting a single agent to operate based on the task, current context, and expected agent performance. Further, the manager role will allow delegations of authority to any agent of the hybrid team and enable intervention when needed. For our framework, we will similarly consider the manager role and aspects of the target scenarios (appropriately modeled) for a manager model which can support diverse contexts and agent abilities. These enable a manager which selects agents for continued operation until an intervention is required. Further, this distinguishes our approach from those highlighted previously.

The need for intervention will be indicated according to constraints provided for the manager. Given constraints, the manager will be cued for an intervention, and subsequent delegation decision, when the constraints are violated. Keeping with the sample driving scenario, the manager could have safety-related constraints such as speed, distance to other vehicles, etc. When a constraint is violated, the manager will be prompted to make a new delegation decision. As we are assuming potentially erroneous agents, the manager is not required to select a different agent at each intervention. This ensures the manager can still choose the agent which best fits the manager's understanding of desirable choices if there is no better alternative. Further, our assumption of erroneous agents means there will often be some likelihood a delegated agent could lead the team to a point where the constraints are violated. The severity of an agent's proclivity for error will therefore impact how suitable they are for a task and how desirable they are for delegation.

To indicate desirable choices, our manager will be trained using Reinforcement Learning and will receive feedback based on its decisions. The feedback, in the form of a reward/penalty, provides a means to learn relative value of choices in the observed world context. Hence, the manager feedback will reinforce our desired balance between the improvement of team performance and the reduction of manager intervention. This is accomplished through a reward function which can indicate positive value for good decisions and penalties for undesirable ones. Further, task-specific values can be associated to manager decisions. These values can be based on the decisions and resulting outcomes to better delineate values. For example, a manager in a driving scenario could place a higher importance on interventions caused by proximity to another vehicle than interventions caused by speeding. The relative values of these intervention types given in the form of penalties would motivate the manager to avoid both these penalties while being willing to accept either to varying degrees. Such levels of relative feedback values allow our representation to support more diverse indications of desirable manager behavior and decisions.

As an added item for manager feedback, we will not make any assumptions regarding a relationship between the manager's feedback and that of the agents. In other words, we do not expect the manager to have access to any information regarding the agents' feedback signals that guide their behavior. Our manager must instead learn purely through external observation and its own feedback. This ensures no unfair assumptions that would allow us to provide unobserved agent information to the manager. For instance, it would be unreasonable to assume the manager could know the exact value a human places on a particular action they would take while accomplishing a task. In the case of driving, such values could be attributed to numerous aspects such as acceleration rates, sharpness of steering angle, proximity to other vehicles, etc. when a human or autonomous driver makes decisions and takes actions. By removing any direct dependence on how agents perceive value, or other key details, our manager can support diverse teams without requiring agent-specific information. On the other hand, we will assume the manager can estimate some aspects of how the agents perceive the world. For instance, we could assume the manager has some model to estimate aspects of what a human or autonomous system would observe in a driving scenario. This could include aspects such as current weather conditions, time of day, sensor/vision occlusions, etc. The manager would not have access to the exact observations but estimates of human or autonomous agent observations could be made based on what the manager can see with respect to context and how the observations are made by an agent. Therefore, we will assume the manager can make its decisions based on its own observations of the environment and some estimate which allows the manager to distinguish between the team agents.

For our team, we assume all agents have already been trained for the tasks they could be expected to perform. This is true for both the human and autonomous agents. Therefore, the manager is not expected to guide any behavior learning for the agents; instead, the manager is specifically tasked with making delegation decisions at critical points. Similar to our assumption of independence of the manager's model of feedback to that of the agents, the assumption of agents which are already trained on the tasks allows us to further isolate the manager from the agents. When the team is not at a critical point, the manager will remain in the background. By reducing the frequency of manager interventions and resulting changes in agent control authority, we reduce the complexity for the team. Since there is need for transferring control when the manager chooses a new agent for delegation, making these decisions more frequently could hinder performance.

In our prior work \cite{fuchs2022cognitive,fuchs2023compensating,fuchs2023optimizing}, we consider related cases of delegation and erroneous agents. While our prior emphasis was on several factors contributing to this work, there are key factors which distinguish our prior efforts from our proposed model. First, in \cite{fuchs2022cognitive}, we consider a mixture of Instance-Based Learning (IBL) \cite{aggarwal2014instance} and Reinforcement Learning (RL) \cite{sutton2018reinforcement} to generate a manager model for a team of two erroneous agents. In this approach, we merged cognitively-inspired models with RL to integrate human-inspired behavior understanding into the manager decisions. The manager was tasked in each step with selecting an authorized agent, based on its cognitive model of behavior understanding, as the team representative. Next, our approach in \cite{fuchs2023compensating} considered the cases of hybrid teams in driving in contexts which impact agent sensing. In this scenario, distinct from our newly proposed model, our manager model made a delegation decision in each state based on direct observations of context and a representation of agent observations. Further, the team observations were provided and merged into a single manager observation. Therefore, our manager provided constant intervention based on access to agent information. 

Our newly proposed model is an extension of our approach in \cite{fuchs2023optimizing}, where we consider an alternative case of delegation in a hybrid teaming scenario. In this work, as in \cite{fuchs2022cognitive,fuchs2023compensating}, the manager made a new delegation decision at the end of each agent's next action. Unlike our work in \cite{fuchs2022cognitive,fuchs2023compensating}, the agents in \cite{fuchs2023optimizing} were not assumed to be operating under the same world model. In this case, we require that all agents operate in the same state space, but we allow for distinct action spaces and subsequent transitions between states. Actions were represented by a composite of 1-3 basic actions (e.g., move left) and completed in a single time step. This representation enabled support for agents which may differ in how they can operate in an environment, but our model relied on their actions being based on the primary base-level actions of the grid environment. By restricting to matching states, but allowing distinct actions/transitions, the manager could support a limited diversity in task execution. On the other hand, the team and manager all relied on consistent representations of desirable behavior by emphasizing shortest paths. In our new model, we instead completely isolate the manager's dependency on the agent models. Our new manager will observe only its own view of the states and observe the resulting change in state caused by the trajectory of a delegated agent. The manager will not observe the intermediary states, actions, or rewards, just the next state where the an intervention occurs. This removes any reliance on consistency between the manager and agent models. Further, our new manager model extends all previous approaches by removing the restriction to a manager delegation at each time step. This ensures support for agent operation across multiple time steps. Additionally, we isolate the manager's notion of desirable behavior from that of the agents. Unlike our previous approaches, we do not assume the manager is motivated by the same notion of desirable behavior as any of the agents.


We summarize the key contributions presented in this paper as follows:
\begin{itemize}
    \vspace{-1mm}
    \item The definition of a \emph{general} framework for an RL-based delegation manager.
    \item A manager model which supports hybrid teams of diverse agents (see Figure~\ref{fig:managed_team_diagram}).
    \begin{itemize}
        \item Support for team members with already learned models of behavior
        \item Reduced manager interaction/interference with team
        \item Increased team performance through optimized team member selection
    \end{itemize}
    \item Model conforming to RL theory converging to optimal solution.
    \vspace{-1mm}
\end{itemize}
To demonstrate the operations of the general framework in a concrete relevant case, we simulated a driving task with a team of human and AI drivers. This scenario demonstrates a key teaming task with a hybrid team of two agents. From our experiments, we observe a positive impact of our manager. In teams with an agent presenting performance significantly higher than the alternative, and with strong or perfect behavior, our manager demonstrates a clear ability to recognize the value of this agent. For teams with a more even level of performance across the agents, and with a mixture of context factors, we see some cases resulting in increased team performance of up to $\approx 187\%$ that of the best solo agent performance.

The remainder of this paper is as follows: Section~\ref{sec:related_work} outlines additional works of note which relate to and inform methods used in our framework; Section~\ref{sec:background_and_framework} provides supporting background material underpinning our proposed framework; 
Section~\ref{sec:driving_behavior} outlines the team agents and their behavior; Section~\ref{sec:results} defines our test scenarios and provides results; and Section~\ref{sec:conclusion} concludes the work.

\section{Related Work}\label{sec:related_work}

Regarding comparisons with existing state of the art (SoA), we surveyed recent and relevant works. According to our search results, we were unable to identify an approach directly comparable to ours. To clarify, we note several recent works which investigate related concepts \cite{Bucinca_Swaroop_Paluch_Murphy_Gajos_2024,Nofshin_Swaroop_Pan_Murphy_Doshi-Velez_2024,Meresht,Richards_Cowell-Butler_2022,Tan_Koleczek_Pradhan_Perello_Chettiar_Ma_Rajaram_Rohra_Srinivasan_Hossain_et}. These works investigate topics of human-AI teams and cases of intervention, but the underlying methods, AI actions, and use cases differ significantly from our model. For instance,  \cite{Tan_Koleczek_Pradhan_Perello_Chettiar_Ma_Rajaram_Rohra_Srinivasan_Hossain_et} investigates a model of interventions by a Reinforcement Learning agent with a goal for minimized interventions. On the other hand, the AI agent augments/extends the abilities of the human rather than serving as an alternate decision-maker. Further, the state space of the AI agent integrates the decisions of the human for concurrent behavior. Similarly, \cite{Nofshin_Swaroop_Pan_Murphy_Doshi-Velez_2024} presents a similar concept of intervention, where the proposed method is intended for helping guide human decisions toward their goal. Related, \cite{Bucinca_Swaroop_Paluch_Murphy_Gajos_2024} presents a method for learning when it is best to intervene in a human decision-making process. In this case, the agent has an action space such that the agent can either offer no assistance, provide a helpful explanation, provide an explanation with a recommendation, or wait for a request from the human for assistance. In this case, the agent is tasked with learning how best to use these specific intervention types to assist the human decision. 
For \cite{Richards_Cowell-Butler_2022}, a representation of tasks and sub-tasks is presented. The representation indicates the potential for diversity in the underlying activities that an agent, human or AI, will perform in pursuit of a goal may differ despite having the same overall result. In other words, the accomplishment of a task, and its underlying components, will depend on a sequence of decisions by the agent, which can differ between humans and AI. This indicates there is the possibility that the sequences of actions/decisions would vary between humans and AI, so the representation of hybrid decisions would need to account for the combination of these sequences to model the team completing a higher-level task. Again, the concepts of this work are closely related, but the work offers no explicit model or approach. This work appears to instead be meant as an investigation into the overall concepts, not as a proposed solution method. Additionally, \cite{Meresht} introduces a 2-Layer MDP model to represent policies in a teaming case. The authors consider agent costs, delegation switching costs, and a cost for actions. In \cite{Meresht}, the authors are providing a method more closely related to our previous scenarios with delegations per state/time step, which we have removed from our model. Further, the model is still based on all agents operating with the same state space and action spaces. Therefore, with respect to the existing SoA, to the best of our knowledge we present a novel approach to managing hybrid teams of diverse (erroneous) agents across multiple decisions and time steps. Therefore, a direct quantitative comparison with alternative approaches is not possible and therefore we compare different cases where our approach is or is not in operation, or where a naive solution (random manager) is used.

\subsection{Hybrid Reinforcement Learning}

Our management of a hybrid team of agents shares common notions found in some Hierarchical Reinforcement Learning (HRL) and related methods. Commonly, these methods will attempt training models with a hierarchical representation. At a lower level, sequences of choices by a behavior model generate sequences of outcomes until the sequence terminates. At a higher level, these sequences and behavior models can be represented as a single ``option,'' which can be selected by a behavior model at the higher level. As an example, the approach outlined in \cite{garcia2020learning} demonstrates a method which attempts to learn ``reusable'' options. In this paradigm, the agent reuses and learns from past experiences to add new options when there is an indication it can improve performance. For another example, \cite{erskine2022developing} demonstrates an approach which includes additional consideration for the compatibility of successive options. The resulting Cooperative Consecutive Policies (CCP) approach learns multiple option policies which are used to accomplish a larger task while reducing the complexity of switching from one option to the next. In the context of multi-agent operation, we see extensions of the HRL concept to consider additional agents in the same environment \cite{chakravorty2019option,chen2022multi,kurzer2018decentralized,rohanimanesh2002learning,singh2020hierarchical,yang2022ldsa}. 
In the multi-agent case, there is a group of two or more agents operating in an environment concurrently, so the agents can be operating simultaneously to accomplish some number of tasks. These can range from cooperative or compatible tasks, as seen in \cite{yang2022ldsa}, where knowledge can be shared to help solve compatible tasks, and compatibility of agent to task is considered to ensure the best use of agents for the given tasks.

The approaches in these topics are indeed relevant to our approach but include some key differences. For instance, we are starting from an assumption of existing policies which can be used to accomplish a task. In other words, unlike HRL and related topics, we are not trying to both learn how to accomplish and learning when to utilize a particular policy. Our manager will instead be tasked when recognizing the best utilization of the team members at its disposal, which reduces the problem complexity as we do not need to account for training multiple behavior models simultaneously. This would be akin to an options framework with fixed option policies. The elimination of the need to train the manager and team members ensures we support scenarios where the team has agents with already trained behavior models. This allows our approach to support the combination of humans and autonomous systems which have already learned their own method of attempting a task. Further, despite our manager operating with hybrid teams including two or more agents, we are not considering cases of concurrent actions. Our team paradigm is one in which only one agent is responsible for the team's current operation. As such, our manager is not operating in a multi-agent setting; instead, the manager is observing a sequence of single-agent operations. Therefore, our management approach is not related to this multi-agent setting. Given this distinction, we make our comparisons to the more closely aligned single-agent models.




\subsection{Delegation}

A distinct but related scenario is that of delegation \cite{candrian2022rise,fuchs2022cognitive,fuchs2023compensating,fuchs2023optimizing,jacq2022lazy,balazadeh2020switch,straitouri2021triage}. 
In delegation models, the manager is again tasked with the selection of an operating agent given a current state of the team. The approach in many delegation cases is to consider a measure of performance and/or cost with respect to the team of agents. The manager is then responsible for recognizing which agent's behavior is best for a situation while also avoiding unnecessary costs incurred by delegating authority to a particular agent. In such an approach, as opposed to our framework, it would be fair to characterize the manager as much more deeply involved in the team's progress. The manager would require mechanisms to more closely model performance and costs associated with the delegation choices, so there is a need for representations of these corresponding values (e.g., model of agent rewards, agent operational costs, etc.). Hence, the manager will require a method by which it can make these estimates and use them when learning associations between delegations and outcomes. For instance, \cite{straitouri2021triage} provides an approach to define and train a Reinforcement Learning model which can operate under algorithmic triage, with algorithmic triage referring to operating in a manner where tasks are shared/delegated according to a prescribed automation level. Based on the accepted level of autonomy afforded to the autonomous agent(s), and related performance/cost factors, the decisions to delegate can be constrained/optimized. In \cite{balazadeh2020switch}, various costs are associated with several components of the scenario (e.g., cost of delegation change, agent usage cost, etc.) which impact the overall reward of the manager. Therefore, both agent cost and their performance will impact the manager reward. Such a consideration can also be seen \cite{jacq2022lazy} which demonstrates bias toward a lower-cost (or ``lazy'') policy when possible.

For our approach, we are primarily concerned with a manager which can delegate authority according to a more general model of selection. By assuming a model of costs and performance, the referenced approaches are indicating assumptions regarding the manager's access to information to measure these. In our case, we remove this assumption, so the manager is instead relying on its own independent sense of acceptable behavior (and no agent cost). More specifically, the manager should be governed by a higher-level measure of acceptable behavior which is independent of the team member currently taking actions. The manager will instead be concerned with ensuring the team can succeed in its task with a minimized need for interventions from the manager. This is to say that the manager is focusing on utilizing any agent which can succeed with the minimal amount of constraint violations. Hence, the granular measure of cost or behavior of an agent is unnecessary so long as they fit within the confines of acceptable behavior.

\subsection{Perception and Reasoning Failures}

An additional aspect of our use of the driving task is the notion of perception and reasoning failures. Humans and autonomous systems which rely on sensing to understand their environment are inherently reliant on the accuracy of their measurements when it comes to making good decisions. If there is a failure in accurately perceiving the environment, the resulting decision will be based on incorrect information. Depending on the scenario, this could result in catastrophic failures, such as collisions, which could have been avoided via a more accurate understanding of the environment. In this regard, \cite{badrloo2022image, guo2019safe, rosique2019systematic, zhang2023perception, zimmermann2020adaptive} demonstrate multiple aspects of the environment which can adversely affect the perception and reasoning accuracy (e.g., weather, sensor type, etc.). For both humans and autonomous systems, vision-based observations can be altered by something as simple as snow/ice obstructing the sensor or human's vision \cite{secci2020failures}.

Another factor in the context of sensing and reasoning is the ability to recognize the significance of an observation. In autonomous systems, a myriad of issues can lead to failed sensing. The perception can be altered via attacks such as small perturbations 
\cite{akhtar2021advances, cao2022emerging, deng2020analysis, eykholt2018robust, zhang2023perception} leading to drastically changed interpretations/classifications. These attacks can often be achieved by something as simple as a sticker on a stop sign \cite{eykholt2018robust}, which could obviously lead to severe consequences. Another critical component of computer-vision in the context of sensing and perception is segmentation. In the case of a segmentation failure, the autonomous system may fail to recognize the existence or significance of an entity in the environment \cite{zhou2019automated}. Again, this can result in failed object detection \cite{rahman2019did} and catastrophic failures.

Given our hybrid team scenario, it is paramount that the manager can learn the best delegations when interventions occur. Further, the manager being given feedback which discourages frequent interventions provides incentive to identify agents which can succeed and reduce the risk of future intervention. Implicit then is the need to help the team overcome potential deficiencies any agent may have. If a human's perception and reasoning is inhibited by contextual factors such as adverse weather, low light, etc., then a manager should recognize the likely reduction in human performance. Similarly, autonomous systems have demonstrated biases and susceptibility to deception which could likewise lead to reduced performance. Therefore, our manager may oversee operation of a team where some/all of the agents are operating with impacted sensing or reasoning abilities. Regardless, the manager should still learn suitable delegation decisions at critical points.

\section{The framework}\label{sec:background_and_framework}

In our framework, the behavior models are learned through interactions with an environment. The operation of agents will occur in an environment represented by states which indicate the context of the team at the current time. The states, actions, and subsequent transitions between states is represented by a Markov Decision process (see \autoref{sec:mdp}). This allows our models to be learned following conventions of Reinforcement Learning (RL). Using concepts of RL allow us to both make observations and select actions in an environment while also optimizing the association between states, actions, and outcomes.

\subsection{Motivation and Background}

For our manager, we are optimizing its delegation decisions and minimizing the need for manager interventions. With respect to an MDP representation, we need to link the observations a manager will make following each delegation interval. As the manager is not making any decisions until a new intervention, the states in which a manager acts are reduced to only those in which an intervention is required. We therefore can split the state representation into those states which require intervention and those which do not. The operation of the delegated agent will continue in all states which do not require intervention and will terminate in a state which does require intervention. This model of operation can be linked to the concept of Absorbing Markov Chains (AMC) \cite{morris1968solving}. In the AMC model, states which terminate an episode or only make transitions to themselves (i.e., loops) are considered absorbing and so the agent is unable to leave them. For our manager at an intervention state, the team can be considered as stuck in an absorbing state until the manager makes its new delegation. The new delegation generates a new AMC which starts from the team's current state and terminates in any reachable intervention states.

With observations of team performance and the feedback resulting from decisions, our manager will learn a behavior policy $\pi_m$ to minimize the likelihood of undesirable team behavior and need for manager interventions. More formally, for team $D$ and environment $E$, the goal of the manager is to learn an optimal behavior model $\pi_m$ to optimize
\begin{equation}
    \min_{\pi_m}P(f|D, E),\qquad\min_{\pi_m}\sum_{s_i\in\{s_1,\dots,s_m\}}\ind{\iota|s_i}
\end{equation}
where $f$ indicates a team failure/error, team path $\{s_1,\dots,s_m\}$, and $\iota$ indicating a manager intervention resulting from behavior violating manager constraints. Therefore, given a team of possible agents ($D$) and the context in which they operate ($E$), the policy we seek ($\pi_m$) should jointly minimize the probability of mistakes ($P(f)$) and the number of interventions of the manager. As we cannot model directly $P(f|D, E)$, we utilize Reinforcement Learning so our manager can learn an optimal behavior model through observations of team performance and corresponding feedback. As such, we present a manager model to support diverse hybrid teams with a primary goal of minimizing the impact of suboptimal agent behavior and minimizing the manager's need to intervene. Such an approach allows us to manage teams where a task is expected to be performed by a single agent, but with the need to identify which member would best enable team success. These decisions do not need to be permanent, so our manager is expected to maintain awareness of the team behavior/performance to continue making effective decisions.

\subsection{Manager model}\label{sec:manager_model}

For our delegation case, we will not require that the manager serves as a teacher for the agents. While a worthwhile means of dictating team behavior, our intention is to instead assume a reduction of interference from the manager while affording more responsibility to the acting agents. In our framework, the teams of agents are instead expected to have prior knowledge of the task. By prior knowledge, we refer to the experience gained when an agent learned to accomplish the task while training/operating independently. Therefore, each agent will join a team and provide its own behavior model gained through this prior training. Hence, our manager is operating with teams of agents with previously trained behavior models, though skill levels may vary, which means the manager is instead learning to best reduce its need to be involved in the team operations. This is accomplished by having the manager learn to optimize the team behavior with respect to a balance between desirability of outcome and the number of times the manager needs to intervene. This results in a trade-off between the performance of team and the frequency of manager interventions. Therefore, our manager will not be required to reach the maximal team reward if it requires an untenable level of intervention.

To model the scenario for the manager paradigm, we extend the MDP model. Diverging from how the team of agents will interact with the world, our manager will only consider three types of states. These three state types indicate the only key cases in which the manager will make an observation. The first two are endpoints of a trajectory, start and goal, which clearly represent the first delegation and final outcome for the team. The remaining case is the intervention states where a delegation is made. Therefore, by design, the observations of the manager will not include any intermediary states reached by the team. In most cases, this will lead to a reduction in the number of states a manager will observe when compared to the hybrid team, reducing the size of the exploration space for learning a manager behavior policy and simplifying the learning task. Additionally, by eliminating the observation and reliance on intermediary states, we remove the need to factor the agent rewards into the manager model. In our model, the manager will instead learn based on feedback entirely independent of the agent reward to prevent any reliance on information we cannot assume the manager can access. This enables a representation which does not rely on any abstraction which merges multiple behaviors into single collections of actions for reduced complexity when learning behavior. Therefore, to model this team paradigm, the manager models its world via a modified MDP which we define as an Intervening MDP (IMDP) $M_I = \langle S=S_\BR\bigcup S_\BQ, A_M, R_M, T_M, D, \beta, \gamma\rangle$ with
\begin{itemize}
    \item $S$, the MDP state space
    \item $S_\BR\subseteq S$, the set of intervention states
    \item $S_\BQ\subseteq S$, the set of unobserved delegated operation states
    \item $A_M = \{i\in D\}$, the manager action space denoting team agents $i\in D$
    \item $R_M$, the manager's reward function
    \item $T_M$, the manager's transition function
    \item $D$, the set of agents available for delegation
    \item $\beta$, the intervention cue function
    \item $\gamma$, the discount parameter
\end{itemize}
In an MDP, the agent observes its status in the environment in the form of states $s\in S$, and move to a new state by taking an action $a\in A_M$. The probability of transitioning from one state to another, by taking an action, is provided by $T_M$. To indicate how desirable a choice is, the reward function $R_M$ provides feedback. Finally, the discount parameter $\gamma\in[0, 1]$ indicates the importance placed on reaching another state and, implicitly, the states reachable in the future. In our model, we extend the MDP representation to include our support of hybrid teams with manager intervention.

In the following, we will describe the extended MDP aspects to integrate the requisite concepts to include the dynamics of the authorized agents and intervention model. First, recognition of when to intervene is based on constraints. The failure of a constraint is then indicated by the manager's intervention function $\beta(s): S\rightarrow\{0,1\}$. This functions allows the alignment of constraints with the problem to indicate behavior or states which shift the team out of the given limits. For example, a likely use of $\beta$ would be to cue an intervention whenever a safety measure indicates an unsafe state (e.g., driving too fast). On the other hand, when an agent is operating within the confines of the manager's constraints, there is no need for an intervention. Without a need to intervene, by design, the manager continues to wait until a new decision is required. Hence, the $\beta$ function will serve as an immediate signal to the manager that something is wrong, and intervention is required. Therefore, $\beta$ operates as an indicator function given state $s\in S$ such that
\begin{equation}
    \beta(s) = \begin{cases}
        1, \quad \textrm{$s$ violates manager constraints $\vee$ $s=s_t$}\\
        0, \quad \mathrm{otherwise}
    \end{cases}
\end{equation}
with $\beta(s)=1$ indicating the need for intervention and $s_t$ representing a terminal state. The inclusion of terminal states ensures the manager makes observations when a team reaches the end of an episode to complete the chain of observations for the manager.

Under the constraints dictated by $\beta$, the manager dynamics will therefore depend on the policy of the delegated agent. To model the transitions between intervention states, we model the likelihood of reaching an intervention state in a manner relating to Absorbing Markov Chains (AMC). Given a Markov chain with states $s\in S$, the chain is considered absorbing if: 1) there is at least one state which affords only recurrent transitions (i.e. termination or self-transition); 2) all non-absorbing states can lead to an absorbing state in one or more steps \cite{morris1968solving}.

Given these properties, it can be shown that the probability of transitioning to an absorbing state $r\in \SR$ after $n$ steps tends to one as $n\rightarrow\infty$. Therefore, the base transition probabilities can be modeled as
\begin{equation}
    P = \begin{bmatrix}
        \begin{array}{c:c}
            \BI_{k-j} & \BO_{(k-j)\times j} \\
            \hdashline
            \BR_{j\times(k-j)} & \BQ_{j\times j}
        \end{array}
    \end{bmatrix}
\end{equation}
In this model, the upper-left quadrant indicates the transition probabilities from recurrent states. 
As defined, these probabilities are zero for any state $s\neq r$, modeled with an identity matrix with dimension $(k-j)\times(k-j)$, for $k=|S|, j=|\SQ|, \textrm{ and } k-j=|\SR|$. The upper-right quadrant extends this idea to model the single step transitions from recurrent states to transient states. As defined, these transitions must be zero, so they are represented by a matrix of zeros of dimension $(k-j)\times j$. For the remaining two quadrants, these indicate the single step transition likelihoods between transient states and transient-to-recurrent transitions. The inter-transient state transitions are modeled with $\BQ_{j\times j}$ while the transient-recurrent transitions are modeled with $\BR_{j\times(k-j)}$.

With the underlying model of state transitions provided by $P$, define fundamental matrix $N$ of the AMC as $N = \left(I - \BQ\right)^{-1}$. Given fundamental matrix $N$, we can then represent the transitions $s_i\rightarrow s_j$ of arbitrary lengths as
\begin{equation}\label{eqn:amc_transition}
    \{b_{ij}\} = B = NR
\end{equation}
to model the general behavior of the AMC. With such a model, the transition likelihoods are modeled in a time-independent manner. Therefore, the dynamics are represented beyond single time steps and state relationships can be observed.

Extending the AMC concepts, the transitions observed by the manager will correspond to those states which are considered absorbing in the AMC model (i.e., $S_\BR = \{s|\beta(s)=1\}$). 
Hence, we can define the transient states as $S_\BQ=\{s|\beta(s)=0\}$. Again, these would of course be those states in which the constraints are satisfied and the episode has not terminated. With this delineation between absorbing and recurrent states, the manager will make transitions from $s_i\in S_\BQ$ to $s_j\in S_\BR$ according to the policy of the delegated agent $\pi_d$ for $d\in D$ and $\beta$. Hence, with infinite horizon, the manager transitions can be modeled as
\begin{equation}
    P(s_j|s_i,d) = b^{\pi_d}_{ij}
\end{equation}
where $b^{\pi_d}_{ij}$ refers to the corresponding values of Equation~\ref{eqn:amc_transition} under policy $\pi_d$. In other words, the probability of the team reaching a state which cues $\beta$ (i.e., an intervention) is modeled according to the probability of reaching the absorbing state under the selected agent's policy. This is of course dictated by the underlying world dynamics, agent policy, and manager constraints.

With the representation of the manager's notion of performance and definitions of $S_\BQ, S_\BR$, we make a further distinction to indicate the final aspect of the manager's states and transitions. With our use of the AMC-based representation of transition dynamics, we must further extend the notion of transient and absorbing states for our model. As currently defined, the manager will enter an absorbing state whenever the team reaches a state $s$ such that $\beta(s)=1$. Under the standard AMC model, this would mean the team would be in a state with only self-transitions or one which terminates an episode. For our model, there is an extension of this definition so a recurrent state, which is not terminal, will transition to an equivalent state with a new delegation. In other words, the act of making a delegation decision will shift the team from an absorbing state to a starting state in a new AMC. This shift in the connection between AMC models can be represented by extending the $\beta$ function to include an additional flag which indicates whether a delegation was made in the current state. In this case, the model becomes
\begin{equation}
    \beta(s, d_s) = \begin{cases}
        1, \quad (s \textrm{ violates manager constraints} \wedge d_s = 0 ) \vee s=s_t\\
        0, \quad s \textrm{ satisfies manager constraints} \vee d_s = 1
    \end{cases}
\end{equation}
where $d_s=1$ indicates a delegation was made in state $s$ at the current time step.

With this updated model, the new AMC represents a case in which the current state is a new starting state and the delegated agent continues until they either reach a terminal state or a new absorbing state (i.e., $\beta(s_j, d_{s_j}) = 1$). Therefore, the manager is observing the start and end states of a sequence of AMC models. In a given episode, the manager performs an intervention by making a delegation decision whenever its $\beta$ function is cued. Based on the constraints defining $\beta$, we can identify which underlying cause cued the new delegation. Given these cues, and a penalty associated with each, we can define a result-based reward which integrates these cues, their costs, and the value of a particular outcome. Therefore, our manager observes the following reward:
\begin{equation}\label{eqn:manager_reward}
	R=\ind{\textrm{goal reached}} - \tanh\left(\delta\cdot\sum_{x\in\mathcal{E}}\rho_x \cdot c_x\right)
\end{equation}
with cue events $x$ in cue event space $\mathcal{E}$, cue frequency $\rho_x$, cue cost $c_x$, and scaling coefficient $\delta$. The cue frequency $\rho_x$ indicates the ratio of delegation cues of type $x$ which occurred resulting from the team's actions in the episode. The reward is observed at the end of an episode and is then associated with each action of the manager's episode to give an episodic measure of desirable actions. This prevents the need for individually attributing delegations as their significance might be sparse in time. Given this structure for the reward function, we can delineate more clearly between outcome types and their corresponding values by scaling costs. Further, the use of the hyperbolic tangent allows us to bound the penalty while $\delta$ shifts how strongly the penalties degrade the episode's value. More specifically, the hyperbolic tangent enables a reward function which is bounded to $[-1, 1]$, which helps stabilize RL training. Additionally, we can more easily scale the significance of each penalty term by using a $\delta$ term which shifts at what value of $|x|$ and how rapidly $|\tanh(x)|$ approaches 1. These factors allow a smoothly increasing/decreasing measure of value which supports relatively scaled penalties.

With our manager's role in the team, it will not associate all individual team actions with its notion of value. Any state, action, and transition which result in a shift to another acceptable state (i.e., constraint-satisfying state) will be of no consequence to the manager. Therefore, the dynamics according to the perspective of the manager will depend on the currently delegated agent and the manager's decision-making process for interventions. In addition to this notion of dynamics, we restrict the perspective of the manager such that the intermediary states are not considered significant. Therefore, the manager's observations and actions coincide only with states where either an intervention is needed, or the team accomplishes its task. Given our model of states and transitions, the manager policy $\pi_m$ will therefore elicit a value function defined as
\begin{equation}\label{eqn:manager_V}
    V^{\pi_m}(s_j) = \sum_{d\in D}\pi_m(d|s_j)\sum_{s_k\in S_\BR} b^{\pi_d}_{jk}\left[R_M + \gamma V^{\pi_m}(s_k)\right]
\end{equation}
The model in \eqref{eqn:manager_V} enables estimation of value associated with manager behavior in a state. The model estimates the expected rewards (episodic or immediate), given by $R_M$, the manager will observe in a trajectory leading from the current state $s_j$. The values of $R_M$ will indicate the value of each decision, either as an immediate reward or based on episode result, to enable value learning for the manager. In the former case, with respect to \eqref{eqn:manager_V}, the reward would be immediately known when the system enters state $s_k$. In the latter, it would be a delayed reward which is available only at the end of the episode. In this case, the optimization would be done at the end of the episode, by backtracking and assigning values to the explored states, still maintaining the overall compliance with standard RL mechanisms. The estimate is subject to the manager's delegation policy $\pi_m$ and the states which will be observed according to the behavior policy $\pi_d$ of the delegated agent(s). The model can be reduced to a recursive definition which models expected state value according to probability of delegation decision $\pi_m(d|s)$, probability of next observed state for the manager $b^{\pi_d}_{jk}$ (according to the behavior of the delegated agent), and the discounted value of the next observed state $\gamma V^{\pi_m}(s_k)$. As noted in \autoref{sec:mdp}, the discount indicates how much importance is placed on the estimated value of future states. A $\gamma$ closer to zero will result in a strongly myopic view while a $\gamma$ closer to one will indicate more confidence/importance of the next state estimate.

Based on our representation of the manager model, we see two key changes we need to account for over a standard RL model: 1) the policy of a delegated agent; 2) unobserved intermediary transitions resulting from delegated agent behavior. From Equation~\ref{eqn:manager_V}, and our consideration of these key aspects, we can define the optimal value function which represents the largest possible value attainable from a state. This represents a model of the best outcome from an optimal policy. Hence, for our manager model, the representation indicates the optimal value associated best selections for delegation
\begin{equation}
    V^*(s_j) = \max_d\sum_{s_k\in S_\BR}b^{\pi_d}_{jk}\left[R_M + \gamma V^*(s_k)\right].
\label{eqn:optimal_V}
\end{equation}

With the above definitions, we can show our model conforms to the requirements for convergence in Reinforcement Learning (see \autoref{appendix:q_convergence}). Therefore, we can use standard RL techniques to estimate an optimal manager policy $\pi_m$. Consequently, we can optimize the choice of the delegation agent $d$ that maximizes Equation~\ref{eqn:manager_V}. This optimized choice will maximize the expected reward achieved by the manager and optimize the behavior of the entire system.

\section{Framework customization for hybrid driving}\label{sec:driving_behavior}

We now consider a specific instance of our framework, cast to the case of hybrid driving. This is the case of a vehicle that can be under the control of either a human driver (human agent) or an AI agent. The driving team is tasked with successfully moving from a start state to a goal state in a four-way intersection environment (see Figure~\ref{fig:sensing_no_context}). The intersection is without traffic signals, so agents should slow to avoid collisions with others. To simulate driving with traffic, the environment includes the team's vehicle along with two additional background vehicles. The background vehicles operate with the same expectation of traversing the environment until they reach their corresponding goal. To ensure the team encounters critical decision points, the background vehicles will not attempt to avoid the team's vehicle except for cases where the team vehicle is leading a background vehicle. As such, the team must select actions to avoid collisions with all background vehicles.


The team will represent a mixture of behavior models to generate a team comprised of human-like and autonomous systems. We will represent our agents with models akin to rule-based methods. The agents will select acceleration values which maintain a speed appropriate for the current road segment as well as any safety concerns. Regarding safety, the agents must avoid speeds and paths which take them too close to another vehicle which shares or intersects their path. These features provide a context where we can measure driver proficiency and provide a meaningful task for a manager. To generate a diverse set of driving agents, we will augment the perception abilities of the drivers based on contexts. The contexts will provide methods by which we can degrade the detection capabilities of the agents and induce erroneous behavior. We will represent degraded sensing via multiple methods which can obstruct or degrade the sensing of the agents. These methods serve to represent scenarios we could expect in real driving scenarios such as blind spots, weather, etc. Context severity will dictate the performance level of the agent and help motivate the use of a manager to overcome agent shortcomings.

\begin{figure}[hb]
    \centering
    \begin{subfigure}[t]{0.43\textwidth}
        \centering
        \includegraphics[width=0.8\textwidth]{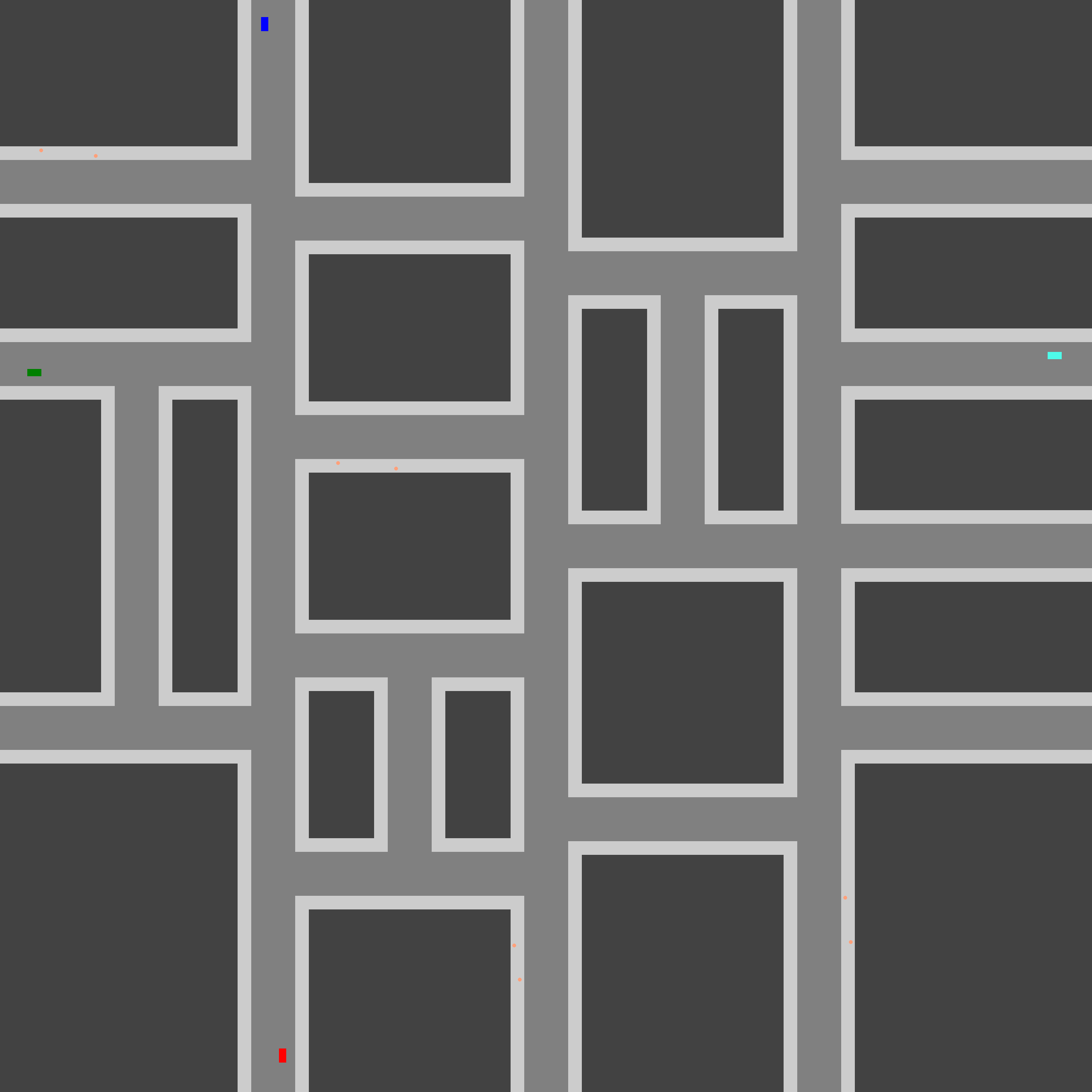}
        \caption{Main environment}
        \label{fig:sample_env}
    \end{subfigure}
    \hspace{3mm}
    \begin{subfigure}[t]{0.43\textwidth}
        \centering
        \includegraphics[width=0.84\textwidth]{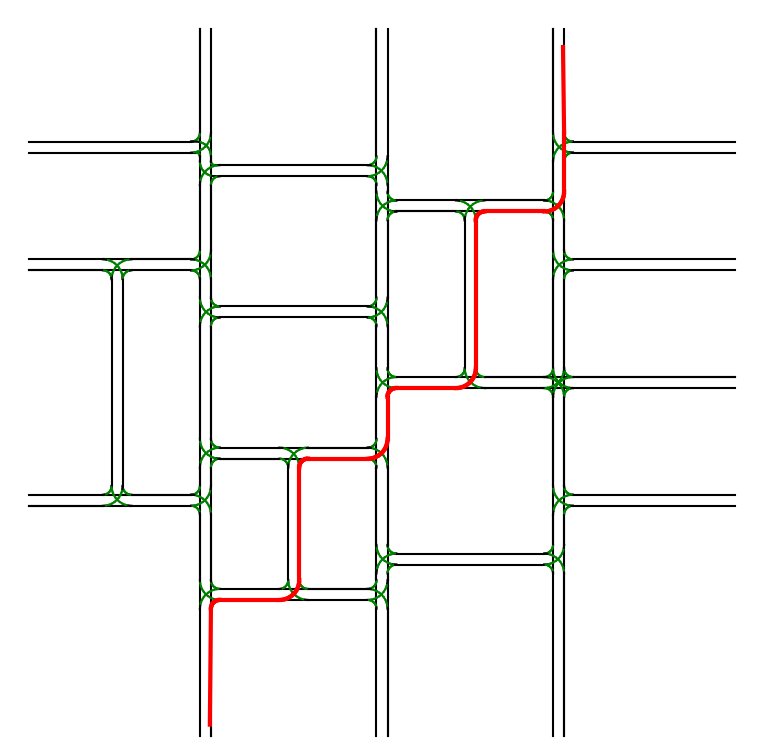}
        \caption{Sample path}
        \label{fig:sample_env_path}
    \end{subfigure}
    \caption{Sample Driving environment (including rectangular buildings, sidewalks, and cars) with sample path.}
    \label{fig:sample_env_and_path}
\end{figure}

We simulate the driving task in a 2D driving simulation based on the CARLO simulator \cite{cao2020reinforcement}. CARLO is a reference simulation environment in the automatic control community. We selected it due to its ability to generate more realistic cases for demonstrating our framework while reducing complexity with respect to aspects we are not emphasizing. More specifically, the CARLO simulator enables vehicle control with estimated kinematics while removing the high-fidelity visualizations/maps not within our representation's scope. In the CARLO environment, cars are represented by rectangles which are controlled via acceleration and steering commands (see Figure~\ref{fig:sample_env_and_path}). Further, the environment supports indications of collisions, which can be used to indicate a driving failure. We augment the CARLO environment to support our scenario as follows. First, we introduce driving paths the cars will follow in an episode. These driving paths only dictate which points in the $xy$-plane are reachable in an episode. The driving agent is therefore allowed to select the acceleration actions it will use to move between states. The selection of acceleration actions will dictate the vehicle speed and its position relative to the other vehicles. For instance, a vehicle will need to select appropriate acceleration actions to maintain speed when it is following another vehicle. Using these paths, we can generate scenarios which result in interactions between the vehicles such as crossing or merging paths.

To create our vehicle paths, we extended the CARLO simulator to generate building blocks for our environments. At the base level, we added support for building environments as combinations of several primitive road types. We support straight lane segments, T-intersections, and four-way intersections. These primitive road types are created in a square cell which can be connected to a compatible neighboring cell. Compatibility is determined by the existence of matching lanes for both cells at their adjacent edge. This enables the connection of an outgoing lane of a T-intersection to an incoming lane of a straight road segment. From the basic road types, we can then extract vehicle lanes the cars can traverse. By limiting the vehicles to these lanes, we can eliminate driving issues resulting from a failure to properly steer the vehicle. In our scenarios, we will limit our experiments to cases of only one or two cells, with the primary cell being a four-way intersection. While our environment supports extending to much larger maps, our focus is on the ability for a manager to monitor a team and assist at critical points.

To ensure the manager has more opportunities to observe these critical points, we generated scenarios with a single intersection to create our critical points. The critical points occur when the team reaches a point where there is a neighboring vehicle which must be considered to avoid a negative outcome. Since the team can traverse paths through an intersection by going left, right, or crossing the intersection, this single intersection will afford three base cases with the possibility of multiple critical points. Critical points will be reached in several ways. First, when there are vehicle paths which have the vehicles intersecting close enough for collision. Second, when two vehicles merge onto the same path, which requires speed matching for rear-collision avoidance. Last, and most simple, when two vehicles are already on the same path. This last case is a sub-case of the previous. The key distinction between the last two is the fact that the merge case includes the added complexity of ensuring one vehicles yields to the other until they can both safely merge paths. Again, our representation of environments, behaviors, and critical points could be extended to support longer paths, but our primary goal was to demonstrate manager performance for critical points. Our use of a single intersection allows this demonstration with multiple critical points in a smaller domain. On the other hand, our representation of the environment would not restrict the manager from making the same observations at additional points in a larger map.

As a further augmentation, we include the ability to represent several driving contexts which impact perception and reasoning. These augmentations serve to impact the detection performance of the drivers with respect to other vehicles. For instance, driving at night, in foggy conditions, or while distracted could all impact at varying degrees the likelihood of recognizing a vehicle of interest. By vehicle of interest, we refer to any vehicle which the primary car could impact or interact with in such a way as to affect their trajectory. As with driving in the real world, the driver's actions are of significance for neighboring vehicles. The behavior of all drivers will impact the safety of all involved. In the case of detection degradation and failure, the driver would be unable to effectively anticipate the consequences of their actions. As a simple example, failure to detect a vehicle in front could lead to a rear collision. The severity of the perception failure will signify the likelihood of unsafe actions due to missed detection.

\subsection{Driver Behavior}

Regarding the driver's behavior, we utilize a behavior policy which is modeled after a rule-based configuration. First, the driver's sensing parameters determine which vehicles in the environment are detectable under the driving context (e.g., obstructed vision blocking detection). Following the selection of detected vehicles, the model then estimates which vehicles will interact given a predetermined safety constraint. For instance, if the a vehicle detects another car which will merge onto the same path, and violating a minimum safety distance, the vehicle should recognize it as interacting. To select an action, the detected vehicle with earliest time of interaction is considered the key interacting vehicle. To remove the occurrence of vehicles avoiding the other, we do not allow two vehicles to defer to each other. Once a vehicle identifies any potential interacting vehicles, there is a sequential selection of driving action.

The vehicles select their action in increasing order of vehicle number in the simulation. When a vehicle has no interacting vehicles, they select an action which allows them to maintain an acceptable speed for their current road location (i.e., slower when turning that driving straight). In the case where a vehicle has detected an interacting vehicle, they choose an acceleration action which allows them to reduce their speed and avoid unsafe interactions. Accelerations of interacting vehicles are made to ensure the two do not remain in the same interaction case at the following time step. The desired speeds and available actions are listed in Table~\ref{tab:driving_parameters_and_thresholds}. Once all vehicles have made their action choices, the simulation applies all actions and the environment updates. If a collision is detected, the simulation episode terminates. If the episode concludes without collisions, the agents will then reach a predetermined goal state.

\begin{table}[htb]
    \caption{Parameters used to constrain driving behaviors. \label{tab:driving_parameters_and_thresholds}}
    \newcolumntype{C}{>{\centering\arraybackslash}X}
    \begin{minipage}{0.57\textwidth}
        \begin{tabularx}{\textwidth}{cC}
            \toprule
            \textbf{Safety and Driving Parameters} & \textbf{Value}\\
            \midrule
            Minimum Safe Distance & $10m$\\
            Goal Reached Threshold & $1m$\\
            Maximum Driving Speed & $9.5 m/s$\\
            Maximum Left Turn Speed & $3.5 m/s$\\
            Maximum Right Turn Speed & $2.5 m/s$\\
            Time between actions & $ 0.1 s$\\
            \bottomrule
        \end{tabularx}
    \end{minipage}
    \hfill
    \begin{minipage}{0.37\textwidth}
        \begin{tabularx}{\textwidth}{cC}
            \toprule
            \textbf{Driver Actions} & \textbf{Value}\\
            \midrule
            Acceleration & $1.0 m/s^2$\\
            No operation & $0.0 m/s^2$\\
            Deceleration & $-1.8 m/s^2$\\
             & \\
             & \\
             & \\
            \bottomrule
        \end{tabularx}
    \end{minipage}
\end{table}

\subsection{Perception and Contexts}

With the parameters defining driving behaviors, our remaining factor impacting driving is the context. We utilize various contexts to simulate degradation in perception. These include cases such as low light, distraction, failed segmentation, etc. Our use of these contexts are intended to model various conditions which would hinder vehicle detection while driving. For example, we can consider driving at night or in foggy weather. First, there is a degradation of visibility with respect to the distance from the observer. The severity depends on the context (e.g., darkness or fogginess), but clearly the conditions will reduce the driver's ability to recognize the existence of other entities in the environment. Similarly, in artificial perception, the sensing systems must convert the sensor input into a model of the environment. In a simple example, occluded sensor (e.g., ice buildup) could hindered perception. As a result, the model of the world is augmented by any inaccuracies in the sensor output. In both the human and artificial agent cases, these perception failures can lead to driving failures (e.g., collision).

In addition to the context types, a context can demonstrate varying degrees of severity regarding how likely a detection failure is to occur. For instance, a failed segmentation in the autonomous system could lead to an entity going entirely unrecognized, while driving at night impacts the detection based on distance and available light. These demonstrate how detection failures may change based on time, speed, and more. Further, the severity level indicates significance. A small occlusion or minor fog levels might have so low an impact that the agent can operate at the same level as when there is no issue, where increasing the severity will of course have the opposite effect. A larger occlusion, higher severity of fog, darker lighting, etc. will all increase the impact on detection and perception. With the use of these contexts, and by varying their severity, we can generate models of behavior which demonstrate a diverse set of failure instances and susceptibility.

\subsubsection{Decayed Sensing}\label{sec:decayed_sensing}

\begin{figure}[ht]
    \centering
    \begin{subfigure}[t]{0.3\textwidth}
        \centering
        \includegraphics[width=0.75\textwidth]{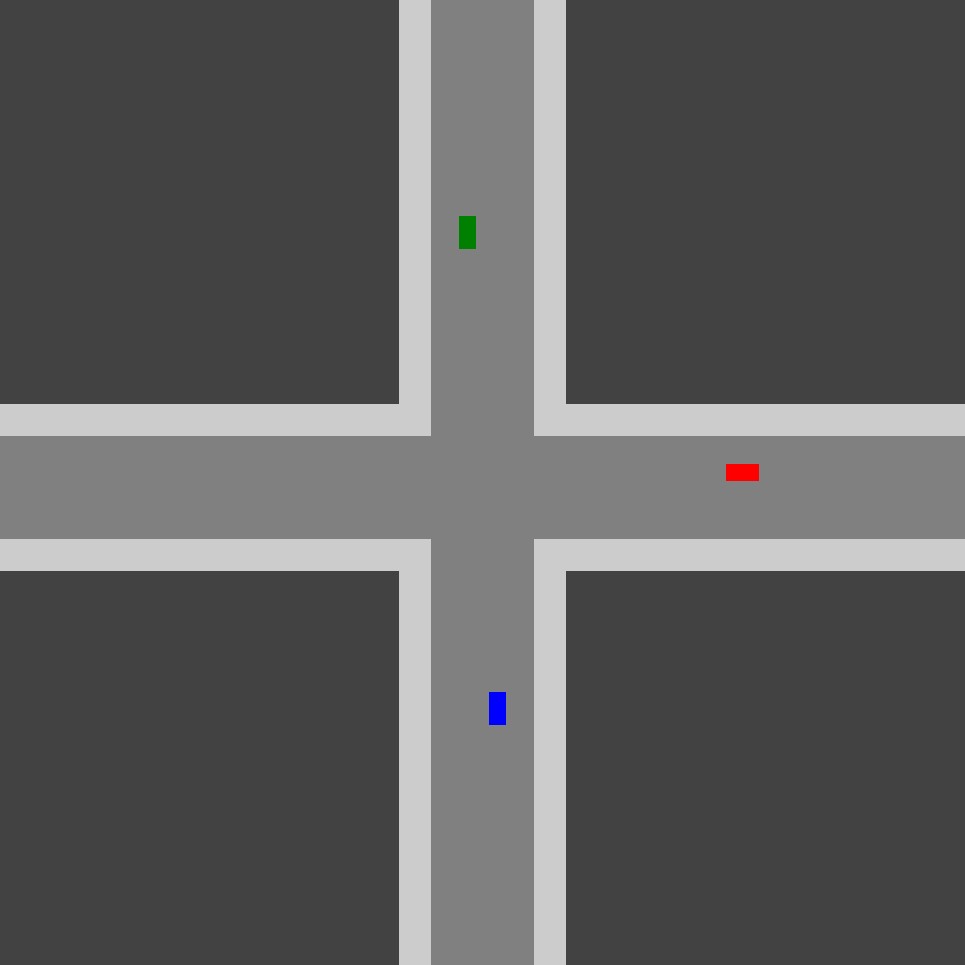}
        \caption{Context-free sensing case}
        \label{fig:sensing_no_context}
    \end{subfigure}
    \hfill
    \begin{subfigure}[t]{0.3\textwidth}
        \centering
        \includegraphics[width=0.75\textwidth]{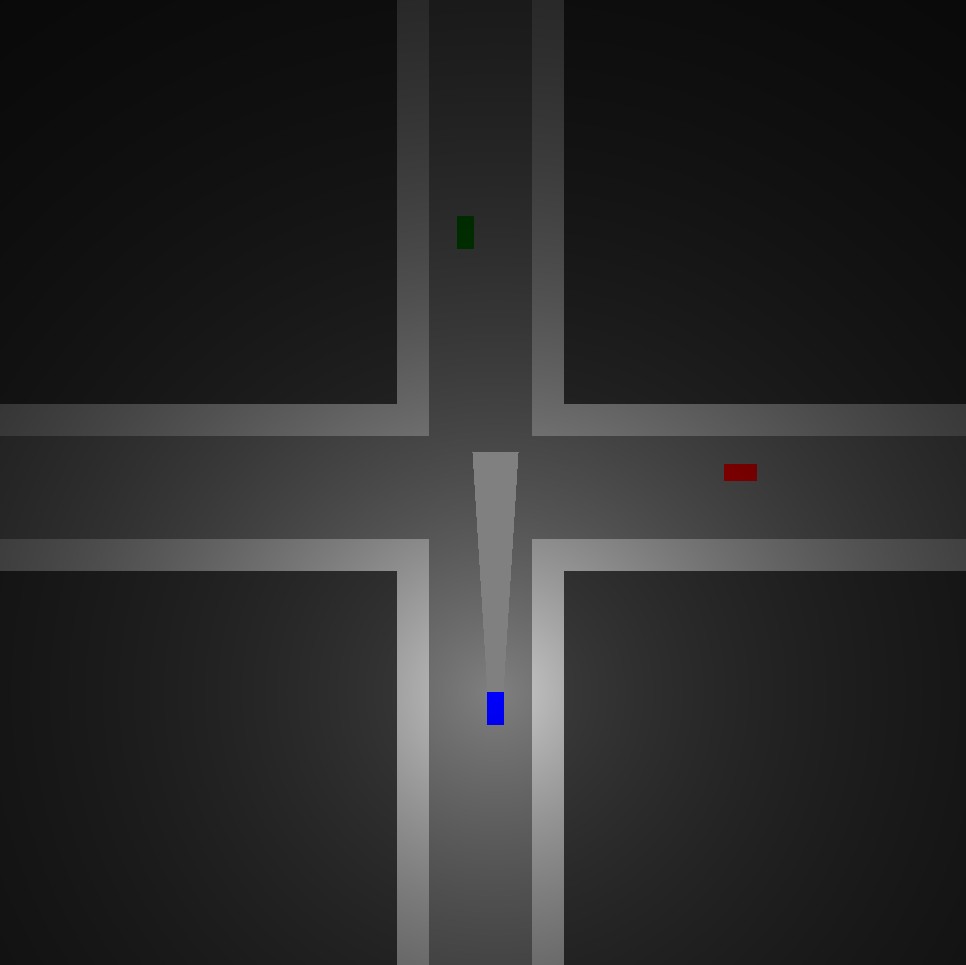}
        \caption{Sensing in low light}
        \label{fig:sensing_night_context}
    \end{subfigure}
    \hfill
    \begin{subfigure}[t]{0.3\textwidth}
        \centering
        \includegraphics[width=0.75\textwidth]{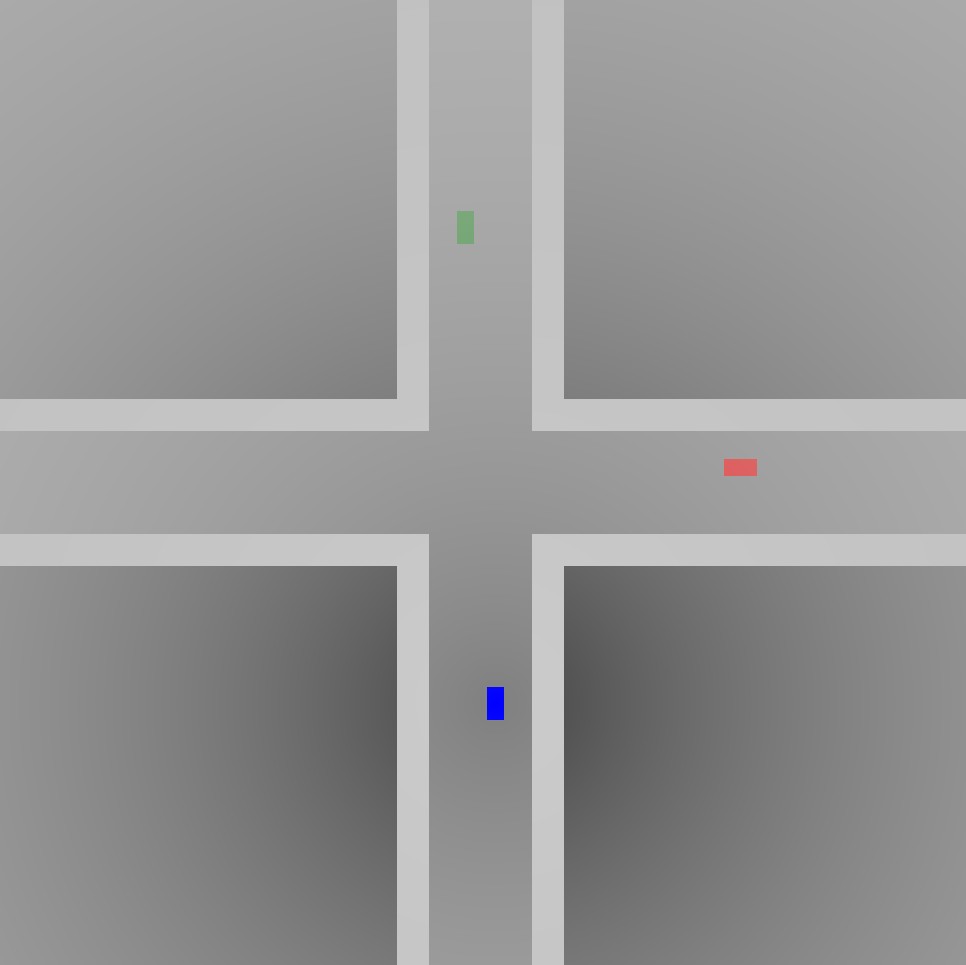}
        \caption{Sensing degraded by distance}
        \label{fig:sensing_fog_context}
    \end{subfigure}
    \caption{Sensing contexts illustrating conditions which may impact the performance/likelihood of detection of other vehicles. Reduced detection increases the likelihood of collision or other driving failures.}
    \label{fig:sensing_contexts}
\end{figure}

In our simulation environment, we demonstrate several contexts exemplified in Figure~\ref{fig:sensing_contexts}. At the most basic level, we include a context which has no impact on sensing beyond merely distance. We allow detection within a prescribed region surrounding the vehicle to imitate natural limitations on vision-based detection. All vehicles outside this region will be considered beyond the sensing range of the driver, while all vehicles inside will be detected with $100\%$ accuracy.
In the first modified context, we simulate night driving by decaying the detection likelihood based on darkness outside the headlight region. The decay is estimated via the following exponential function parameterized to reach a specified decay level at a given distance:
\begin{equation}\label{eqn:exp_decay}
    f(x) = e^{-x/\tau}
\end{equation}
where $\tau$ determines how severely the function decays and $x$ is the distance between the point of observation and the coordinate observed. We generate severity $\tau$ by two parameters $x_\tau$ and $y_\tau$. These denote the distance to the severity point and corresponding the decay value at the severity point, respectively. Hence, define $\tau$ as:
\begin{equation}\label{eqn:alpha}
    \tau = \frac{-x_\tau}{\ln{y_\tau}}
\end{equation}
All distances will then follow this exponential decay to determine the detection likelihood. For night driving, a vehicle in the headlight region has a detection likelihood of $100\%$.

Similar to the night case, we include a case in which detection decays in all directions according to distance. In the human case, this can be treated as driving in foggy weather (see Figure~\ref{fig:sensing_fog_context}). As with the night driving case, the likelihood of detection decays as a function of distance. Unlike the night case, there is no region where the detection likelihood can be restored to the base level. Therefore, this context serves to represent any case in which the sensing ability of the driver should degrade as a direct result of distance regardless of relative direction.

\subsubsection{Failed Perception}\label{sec:masked_sensing}

\begin{figure}[ht]
    \centering
    \begin{subfigure}[t]{0.4\textwidth}
        \centering
        \includegraphics[width=0.56\textwidth]{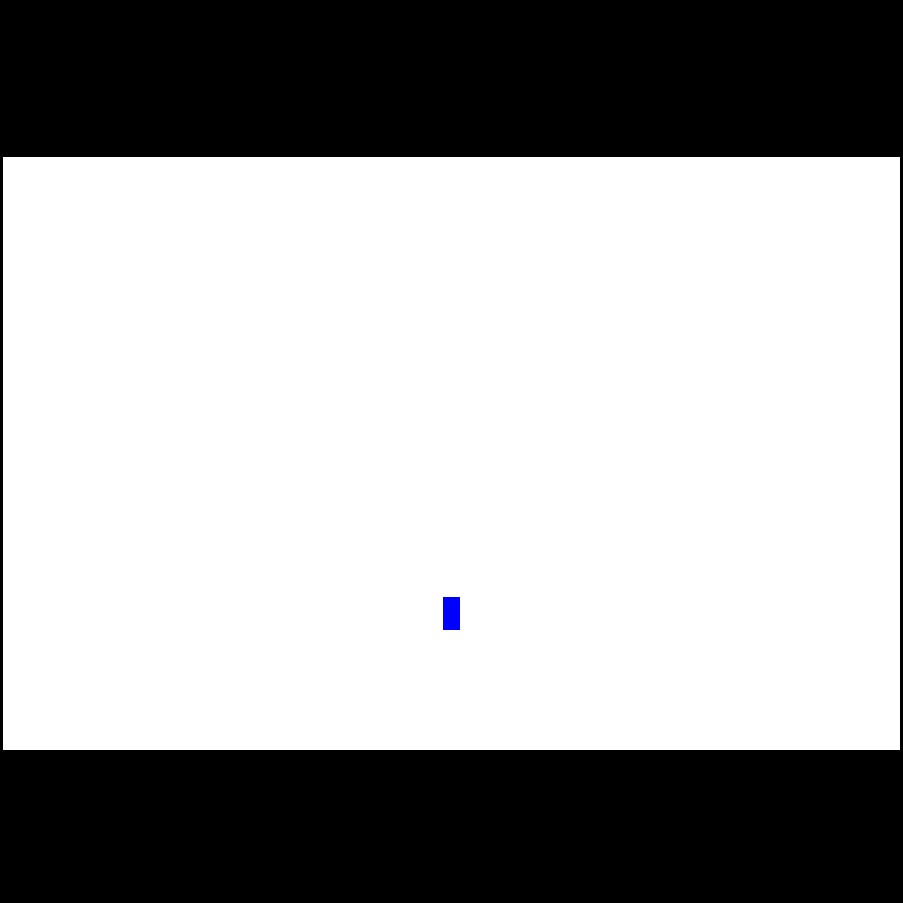}
        \caption{Distracted driver (masked observation)}
        \label{fig:sensing_distracted}
    \end{subfigure}
    \hspace{3mm}
    \begin{subfigure}[t]{.4\textwidth}
        \centering
        \includegraphics[width=0.56\textwidth]{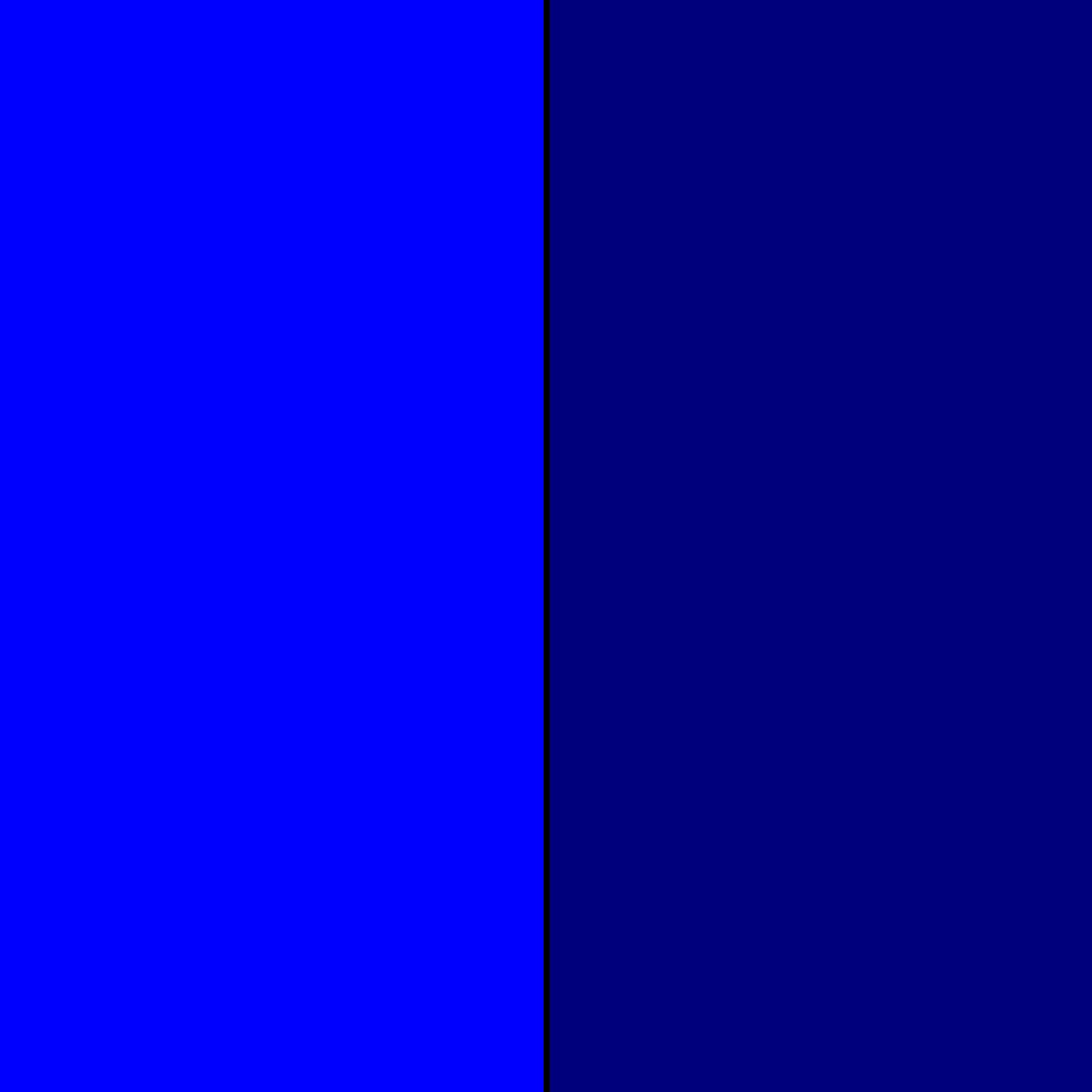}
        \caption{Car color vs Blindness color}
        \label{fig:sensing_colorblind}
    \end{subfigure}
    \caption{Sensing contexts illustrating conditions which may impact the performance/likelihood of detection of other vehicles. Reduced detection increases the likelihood of collision or other driving failures.}
    \label{fig:perception_contexts}
\end{figure}

The final two cases represent another manner of failed detection, where we instead utilize a method based on intrinsic factors. These factors are those such as segmentation fault, obstruction, or distraction, which lead to completely missed detection for a region of the observation space. Regarding the distraction case, the driver is modeled as being unaware of any vehicle in a particular region. This is done by modeling masks which obstruct detection for their region (see Figure~\ref{fig:sensing_distracted}). 
For a reduced severity, these masks can be defined to cover a smaller region of the observation space. When an entity intersects the masked region, the detection likelihood is reduced by the size of the covered area. Based on the resulting observable area of the entity, the agent detection is triggered proportional to the ratio of observed and total entity area. This method is intended to provide a model of obstructed sensing which indicates a connection between severity of obstruction and reduction in detection likelihood.

As another case, we model failed perception of autonomous systems, as seen in works such as \cite{secci2020failures, eykholt2018robust, zhou2019automated, cao2019adversarial}, 
by representing the perception failure is as an inability to recognize a particular vehicle, based on similarity to a blindness color. To measure similarity, we use an approximation via a weighted Euclidean distance as seen in \cite{www_riemersma_2019}. For colors $C_1 = (R_1, G_1, B_1)$ and $C_2 = (R_2, G_2, B_2)$, the color distance is defined as:
\begin{equation}\label{eqn:color_distance}
    d(C_1, C_2) = \sqrt{\left(2 + \frac{\bar{r}}{256}\right)(R_1 - R_2)^2 + 4(G_1 - G_2)^2 + \left(2 + \frac{255 - \bar{r}}{256}\right)(B_1 - B_2)^2}
\end{equation}
where $\bar{r} = \frac{1}{2}(R_1 + R_2)$. As the color distance increases, the likelihood of failure will correspondingly decrease. Therefore, we can model distance-agnostic perception blindness while allowing these failures to still vary by a parameter. In this case, that parameter is the similarity of the blindness color to any of the vehicles.

Given these contexts, we can therefore generate various driver models by selecting driving contexts and varying context parameters to increase/decrease the impact of the context of the driver's perception, which in turn impacts the performance of each member of the team. As such, we can generate multiple teams comprised of agents impacted by different perception failings.

\subsection{Manager Behavior}

To train our manager, we utilize an episodic reward and treat collisions as terminal states. Given the nature of our scenario, with the potential long-term or sparse attribution of actions to outcomes, we utilize an episodic reward for more generalized attribution of action values. This ensures that the manager learns associations between rewards and trajectories rather than specific actions. Over the training process, the manager should therefore learn to prioritize interventions more likely to lead to better trajectories and outcomes. The interventions are dictated by the $\beta$ function, which we base on several factors for the driving scenario. In the context of driving, our manager will seek to maintain the safety of the driving team by including cues related to (i) proximity, (ii) speed, and (iii) collisions. If any of these constraints are violated, $\beta$ will trigger a new intervention or observation of episode termination for the collision case. The values of the constraints for these cues are similar to those used for the driving agents. The manager will be equally concerned with speed relative to the current position on the road as well as the proximity to vehicles which share or intersect the path of the team. Finally, the manager will be most concerned with collisions as these represent a critical failure of the team. Combined, these factors which cue $\beta$ will guide the manager to select agents which can complete the task and minimize the occurrence of behavior which results in constraint failure.


\section{Experiments and Results}\label{sec:results}

In our experiments, we train our manager in a driving scenario for a four-way intersection case. We generate multiple team driving scenarios which include driving tasks requiring successful navigation from a start state before the intersection to a goal state beyond the intersection. The driving tasks are further segmented into left turn, right turn, and driving straight through the intersection. In each case, there are two additional vehicles also navigating the environment that the team must avoid. To increase the problem difficulty and ensure there are critical states requiring effective decisions by our team, the two additional vehicles will not attempt to avoid the team's vehicle when navigating. As an exception, the additional vehicles will avoid a collision with the team's vehicle when following it on the same path, so the vehicles avoid only collisions which the team couldn't be expected to prevent. By shifting the key collision avoidance decisions to the driving team, we ensure the manager must identify which agent(s) are best suited to safely navigate the tasks. The task difficulty is determined by the perception context (e.g., fog) and its corresponding severity. The manager performance is based on how well its selected agents perform with respect to the manager constraints. The manager should learn to select agents which have the best chance of task completion and lowest chance of constraint violation. Further, in the case of unavoidable constraint violation, the manager should learn to prefer agents which will cause the ``least significant'' constraint violation, according to the manager's reward function.

\subsection{Performance Metrics}\label{subsec:performance_metrics}

To measure the manager performance, we measure several aspects of the scenario. First, we track the task completion rate $n_g$ of the managed team.
\begin{equation}
    n_g = \frac{1}{|\mathcal{D}|}\sum_{s_t\in \mathcal{D}}\ind{s_t = s_g}
\end{equation}
for final state $s_t$ from trajectories of experiences in experience set $\mathcal{D}$ and goal state $s_g$. Hence, $n_g$ measures the fraction of success versus attempts. As another measure of performance, we track the frequency of interventions $n_\beta$ in each episode
\begin{equation}
    n_\beta = \sum_{s_t\in\{s_1,\dots,s_m\}} \ind{\beta(s_t, d_{s_t}) = 1}
\end{equation}
with episode trajectory $\{s_1,\dots,s_m\}$. For the presentation of results, we will indicate intervention performance as the mean intervention occurrence $\bar{n}_\beta$ across all test episodes. Combined, $n_g$ and $\bar{n}_\beta$ will indicate how frequently the task is completed and how often the vehicle enters a state which violates the constraints or induces a manager intervention.

To demonstrate the impact on performance from our manager, we will compare against two baselines. The first is represented by the individual agents operating solo in the driving scenario. We use $n_g$ and $\bar{n}_\beta$ to track the individual agent performances in the scenario and context. The use of $\bar{n}_\beta$ in the solo case will not indicate how many times the manager makes a delegation decision but rather the number of times the manager would choose to intervene if the driver were on a team. For the $n_g$ value, this is applicable to all cases as it is a general measure of task completion success rate. The second baseline we use is that of a random manager. In this case, the delegation at each intervention will be made with uniform random selection of the next driving agent. The random manager provides us an indication of both the impact of using any manager and how much improvement is made by learning an optimized manager policy. We will illustrate managed team performance in the following section with plots indicating resulting values for $n_g$ and $\bar{n}_\beta$.

\begin{figure}[htbp]
    \vspace{-3mm}
    \begin{center}
        \begin{forest}
            for tree={
                parent anchor=children,
                child anchor=parent,
                s sep'=10pt,
                l sep'=2mm,
                font=\sffamily,
                },
                before typesetting nodes={
                tempdima/.max={>Ow+P{content}{width("#1")}}{fake=r,leaves},
                for nodewalk={fake=r,leaves}{text width/.register=tempdima, tier=terminus},
            },
            align middle children,
            [$s_i$ 
                [$d_{i,1}$, name=D11
                    [$s_{i+1,1}$ , name=S11_11
                        [$d_{i+1,1}$, name=D11_11_11]
                        [$d_{i+1,m}$, name=D11_11_1M]
                    ]
                    [$s_{i+1,n}$ , name=S11_1N
                        [$d_{i+1,1}$, name=D1N_1N_11]
                        [$d_{i+1,m}$, name=D1N_1N_1M]
                    ]
                ]
                [$d_{i,m}$, name=D1M
                    [$s_{i+1,1}$ , name=S1M_11
                        [$d_{i+1,1}$, name=D1M_11_11]
                        [$d_{i+1,m}$, name=D1M_11_1M]
                    ]
                    [$s_{i+1,n}$ , name=S1M_1N
                        [$d_{i+1,1}$, name=D1M_1N_11]
                        [$d_{i+1,m}$, name=D1M_1N_1M]
                    ]
                ]
            ]
            \draw[dotted] (D11) -- (D1M);
            \draw[dotted] (S11_11) -- (S11_1N);
            \draw[dotted] (S1M_11) -- (S1M_1N);
            \draw[dotted] (D11_11_11) -- (D11_11_1M);
            \draw[dotted] (D1N_1N_11) -- (D1N_1N_1M);
            \draw[dotted] (D1M_11_11) -- (D1M_11_1M);
            \draw[dotted] (D1M_1N_11) -- (D1M_1N_1M);
        \end{forest}
    \end{center}
    \caption{Sample first three branch steps to generate trajectories $\tau_{s_i}\in\mathcal{T}$}
    \label{fig:path_branches}
    \vspace{-0.5mm}
\end{figure}
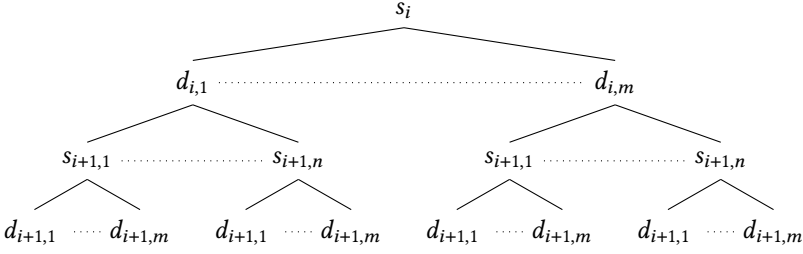

We should note important factors which will influence the demonstrated results. The time between a critical decision and the outcome may span multiple time steps. For example, a driver may fail to slow down to avoid a collision with another vehicle. While the collision is the critical failure, the missed opportunity to slow the vehicle occurs prior to the time of the collision. Therefore, a team's outcome will depend on the combination of manager decisions, when those decisions occur, the criticality of that decision with respect to time to alter courses, decisions of other vehicles, and the behavior of the delegated agent. For episodes with maximum length $H$, we can illustrate an estimate of the impact from these factors by a probability of failure
\begin{equation}\label{eqn:path_failure}
    p(f|s_i) = \sum_{\tau_{s_i}\in\mathcal{T}_{s_i}}\prod_{j=i}^{i+n-1}\pi_m(d_j|s_j)b^{\pi_{d_j}}_{jj+1}\ind{s_{i+n}\in S_F}
\end{equation}
where $f$ indicates a failure and $\tau_{s_i} = (s_i,d_i,\dots,s_{i+n-1},d_{i+n-1},s_{i+n})\in\mathcal{T}_{s_i}$ indicates a trajectory of states $s_j\in\SQ$ and delegated agents $d_j\in D$. The trajectories $\tau_{s_i}$ are those which start in the current state $s_i$ and terminate in a failure state $s_{i+n}\in S_F\subseteq\SQ$. The trajectory length is bounded by the remaining time steps in an episode (i.e., $n\leq H-i$). Equation~\ref{eqn:path_failure} provides a model of trajectory likelihoods given a starting point. These trajectories represent different paths the team would reach if decisions were changed (see tree diagram in Fig.~\ref{fig:path_branches}). By modeling all possible paths which can lead to failure states, we represent the combined likelihood of reaching any failure state. The probability of a path is represented by the product of the probabilities for each connection between nodes (i.e., $\pi_m(d|s_j)$ and $b^{\pi_{d_j}}_{jj+1}$). The sum of all these path probabilities then gives the probability of a failure. Further, we can indicate how a decision earlier in a path impacts the possible later outcomes based on how it impacts the possible subsequent paths.

Keeping with the driving scenario and key time to make a decision, we can see that the likelihood of a failure will depend on whether there are remaining failure states reachable from $s_i$ (i.e., $|S_F| > 0$). A skillful agent will make action selections which reduce the probability of entering a failure state, thereby shrinking $|S_F|$. By reducing the chances of reaching a later failure state, the skillful agent is reducing the size of $S_F$ for subsequent states $s_j$. Therefore, at the next delegation decision, the selection of an erroneous agent will have reduced likelihood of generating a trajectory leading to a state $s_{i+n}\in S_F$. For instance, driving slowly could lead to other vehicles already passing through the intersection prior to your arrival. On the other hand, if another vehicle is approximately the same distance to an intersection and traveling at a similar speed, both you and the other vehicle should arrive at similar times. Therefore, earlier decisions impact what later outcomes are possible, which determines which paths comprise $\mathcal{T}_{s_i}$. Following Fig.~\ref{fig:path_branches}, this would mean eliminating some branches from the tree as they are no longer possible. On the other hand, the selection of an erroneous agent could result in the opposite effect, resulting in more paths to a failure. A selection of an erroneous agent, followed by an erroneous action, will reduce the number of states $s_{i+n}\notin S_F$ the team can reach, increasing the chances of a failure. These factors will be most apparent in the case of a random manager. The difficulty of the underlying task and how frequently the team can encounter these key decision points will impact how the random manager performs. Moreover, the interplay of the manager and agent decisions will result in outcome likelihoods which are not as simple as a Bernoulli trial. Therefore, we cannot expect a random manager uniform policy to generate an approximate success rate of $50\%$. The random manager score will instead rely on how frequently their random choices lead to low values of Equation~\ref{eqn:path_failure}.

For our measure of manager performance via the rate of successful navigation of the team (i.e., $n_g$), the number of test episodes was chosen as $250$. As $n_g\in[0, 1]$, the use of $250$ tests ensures a variance less than $0.07$ for a confidence interval of $95\%$. From these episodes, we also gather our results for $\bar{n}_\beta$. In our scenario, we endeavor for team dynamics which improve the overall performance of the team while also reducing the necessity for manager intervention. Therefore, we feel that tracking the occurrences of delegations to be a valid indicator of team and manager performance. As a final note, we anticipate the delegation frequencies will be lower in the Human-solo and Random cases if the actions of the drivers lead to a collision with another vehicle in the environment. The reduction in delegation frequencies $\bar{n}_\beta$ is a result of the shorter episodes caused by this team failures. In other words, these scenarios result in fewer time steps and consequently fewer opportunities for delegations before episode termination. This of course means that the delegation count is lower, which looks like better performance, but these correspond to cases where the team is unable to complete the task and performs worse than cases with higher $n_g$.

\subsection{Driving with Intervening Manager}\label{sec:manager_parameters}

To motivate learning, the manager observes the cue costs in Table~\ref{tab:manager_rewards}. Note that the manager still receives a cue cost for collisions to further indicate the severity of this undesirable outcomes. This allows for a more severe distinction between sub-optimal team performance and catastrophic failures such as collisions.
\begin{table}[htb] 
    \vspace{-0.5em}
    \caption{Manager Reward Parameters. \label{tab:manager_rewards}}
    \newcolumntype{C}{>{\centering\arraybackslash}X}
    \begin{tabularx}{\textwidth}{CCCCC}
        \toprule
        \textbf{Parameter} & $\rho_{\textrm{velocity violation}}$ & $\rho_{\textrm{proximity violation}}$ & $\rho_{\textrm{collision}}$ & $\delta$\\
        \midrule
        \textbf{Value} & 5 & 20 & 75 & 300\\
        \bottomrule
    \end{tabularx}
    \vspace{-1em}
\end{table}

\subsubsection{DeepRL Model}

To define our manager, we utilize the Soft Actor-Critic (SAC) \cite{haarnoja2018soft} with prioritized experience replay (PER) \cite{schaul2015prioritized}. The manager is trained by first generating behavior samples of each solo agent for a set of episodes followed by a set of random exploration episodes. The number of pre-training and random episodes is set as a hyperparameter for each manager. Following the pre-training and random episodes, the manager learns a behavior policy according to the standard training method. Additionally, the manager performs soft model updates after each episode for each experience tuple which tracks the state transitions and episodic rewards. The SAC model was selected to reduce early bias in the highly sparse driving environments and promote entropy-based exploration.

\subsection{Driving Task}

The team's actions will be indirectly considered with respect to the managers $\beta$ function via monitoring of the resulting state transitions. The manager will not make direct observations of the choices of the delegated agent but will instead make indirect observations based on the updated vehicle state at the next intervention. Therefore, the manager observations will be made in states where the driving team reaches a state which triggers an intervention. The interventions will result in a possible change in delegated authority. The choice of driver will therefore also impact the overall outcome of the team. Finally, the success of the delegated agent will be dictated by the context for the agent and its impact on their sensing.

\subsection{Performance in case of biased teams}

In the following sections, we demonstrate manager performance in teams with a driver which can operate in an error-free manner. The remaining driver is subject to varying degrees of degraded performance in order to indicate how sensitive the learned manager is to these discrepancies. The intention is to illustrate the ability of the manager to recognize when there should be a clear indication of one agent being more desirable. This does not mean that one agent would fail to drive in all states but rather that they are more likely to make a mistake in critical states. We demonstrate performance in the various sensing degradation contexts we describe in Section~\ref{sec:decayed_sensing} and Section~\ref{sec:masked_sensing}. Again, these contexts offer configuration which can alter the severity of the impact on agent perception. As such, we utilize several configurations to generate unique teams.

\subsubsection{Distracted Driving}

As our first scenario, we demonstrate a team with a distracted human driver and an autonomous driver with the base error-free detection context. In this setup, the human driver will fail to recognize the existence of any other vehicles while the autonomous system will detect all vehicles in its range. This is accomplished by masking the observations to eliminate any vehicle detection for the distracted driver (see Figure~\ref{fig:sensing_distracted}). Therefore, the team includes one agent which will fail in all cases while the other will succeed so long as its behavior policy can act according to the detected vehicles. In this set of driver contexts, we would expect our manager to learn a bias towards the autonomous driver as they would be the only agent capable of successfully navigating the environment.

For the random manager, there are several key factors impacting performance. First, as outlined in Equation~\ref{eqn:path_failure}, there will be states where the selection of the distracted human will increase the number of failure states reachable from the current state. On the other hand, we should note there are many states where the actions of the agent are inconsequential with respect to collisions. If the distracted driver is not in a situation with an interacting vehicle and their action is the same as the AI agent, then both will result in the same next state. The second important factor is the specific task for the team. In each of the left, right, and straight driving cases, the movement of the vehicles will result in different critical points. Therefore, the set $S_F$ will differ across each of the three tasks, which could lead to significant variance in the same context between the three tasks. Last, by definition, the random manager will not be motivated to reduce the number of interventions. As such, the variance of the decisions from the random manager between episodes and tasks will result in highly diverse experiences and opportunities for success/failure according to the task. In the three scenarios, the results indicate a higher instance of critical states and more potential failure states in the left turn and right turn tasks than the straight driving task. This is most likely attributed to the fact that the two additional vehicles will slow to avoid each other, which could offer our hybrid team a chance to miss these vehicles by chance as a result of the window of opportunity afforded by the slowing additional vehicles. In this case, the team may pass through the intersection before there is a chance to interact with another vehicle.

\begin{figure}
\vspace{-1.5mm}
    \begin{subfigure}[t]{0.475\textwidth}
        \centering
        \includegraphics[width=0.875\textwidth]{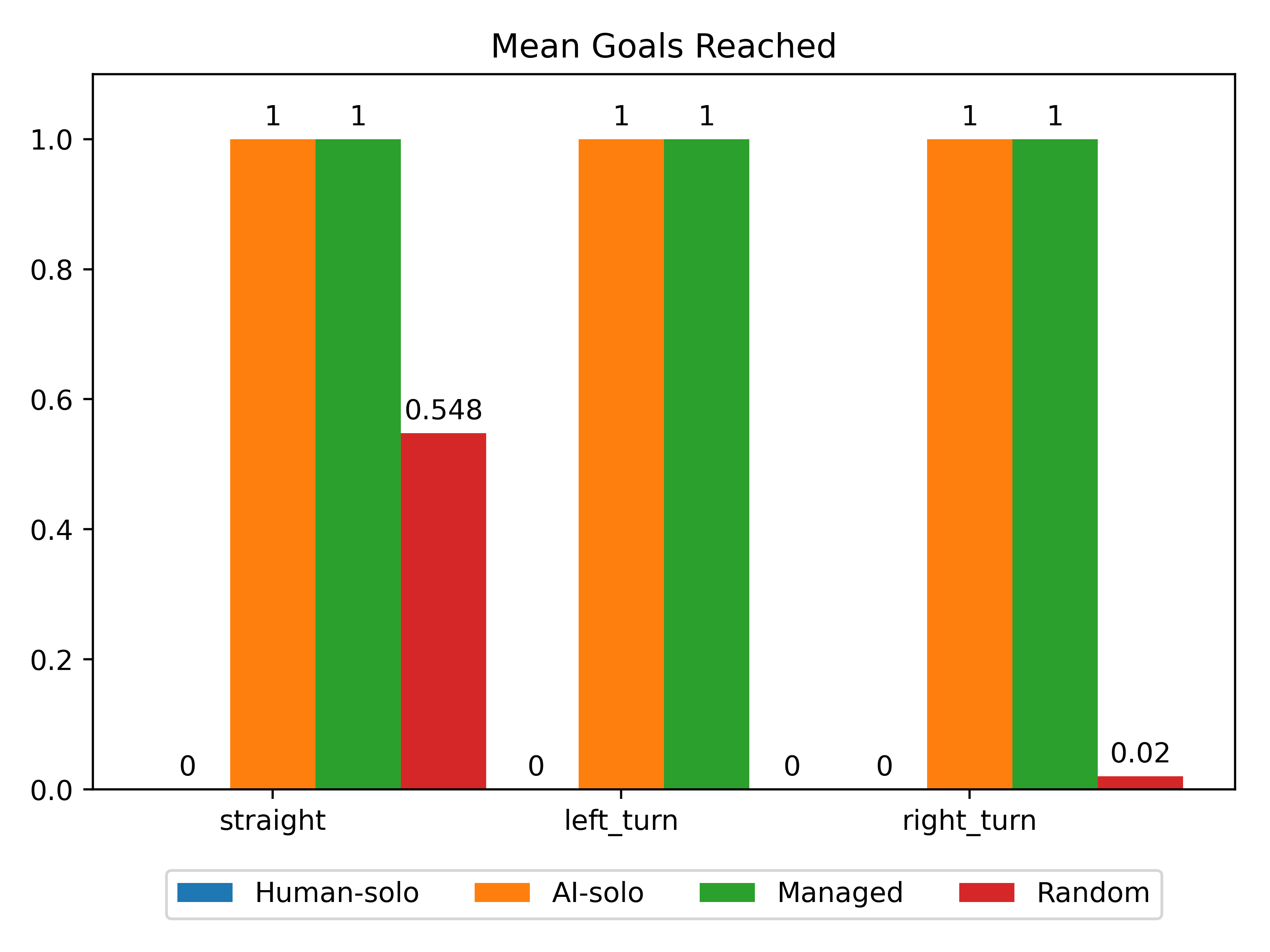}\vspace{-2mm}
        \caption{Distracted Driving}
        \label{fig:distracted_driver}
    \end{subfigure}
    \begin{subfigure}[t]{0.475\textwidth}
        \centering
        \includegraphics[width=0.875\textwidth]{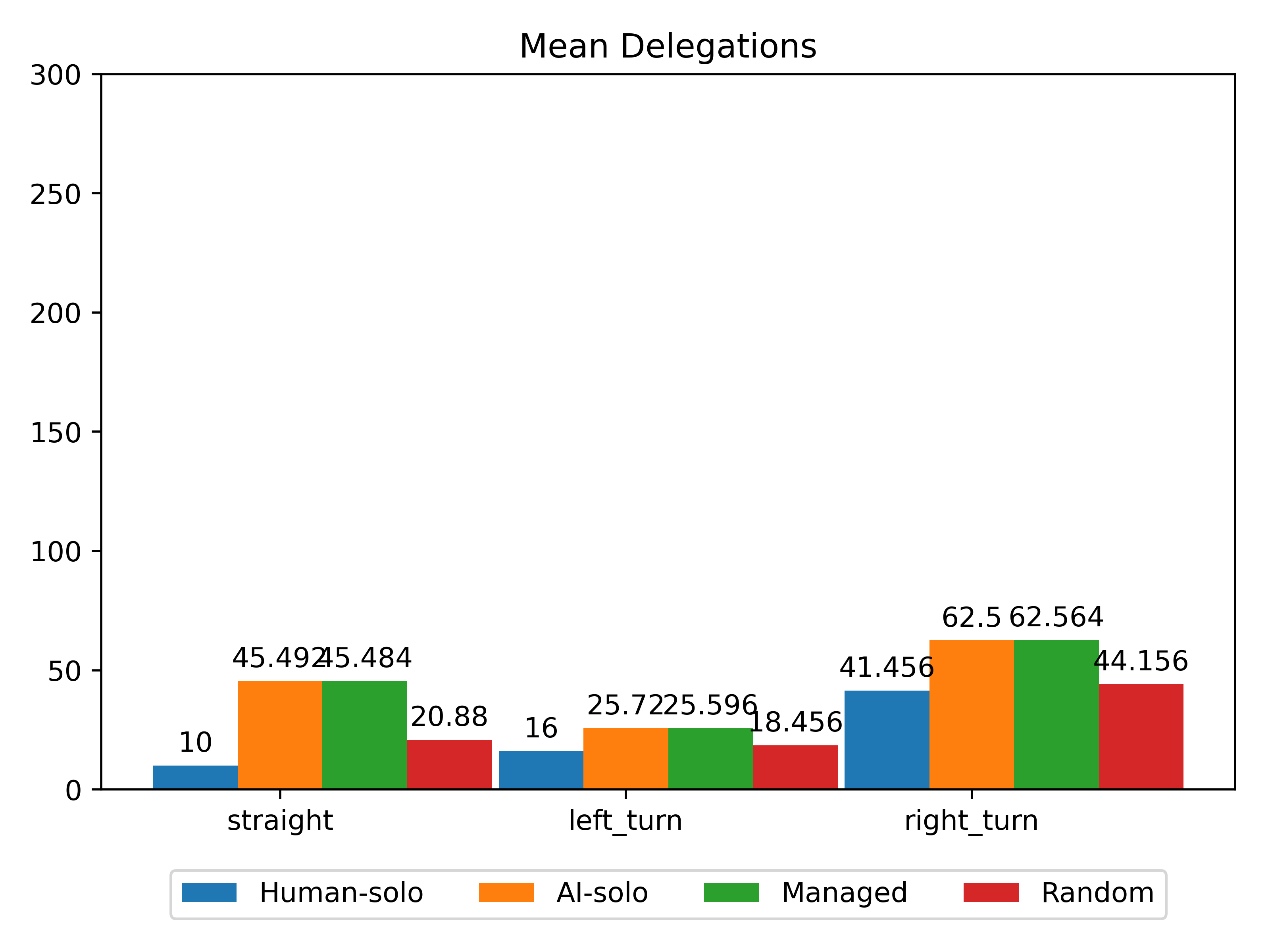}\vspace{-2mm}
        \caption{Distracted Driving}
        \label{fig:distracted_driver_delegations}
    \end{subfigure}\\\vspace{-0.25mm}
    \begin{subfigure}[t]{0.475\textwidth}
        \centering
        \includegraphics[width=0.875\textwidth]{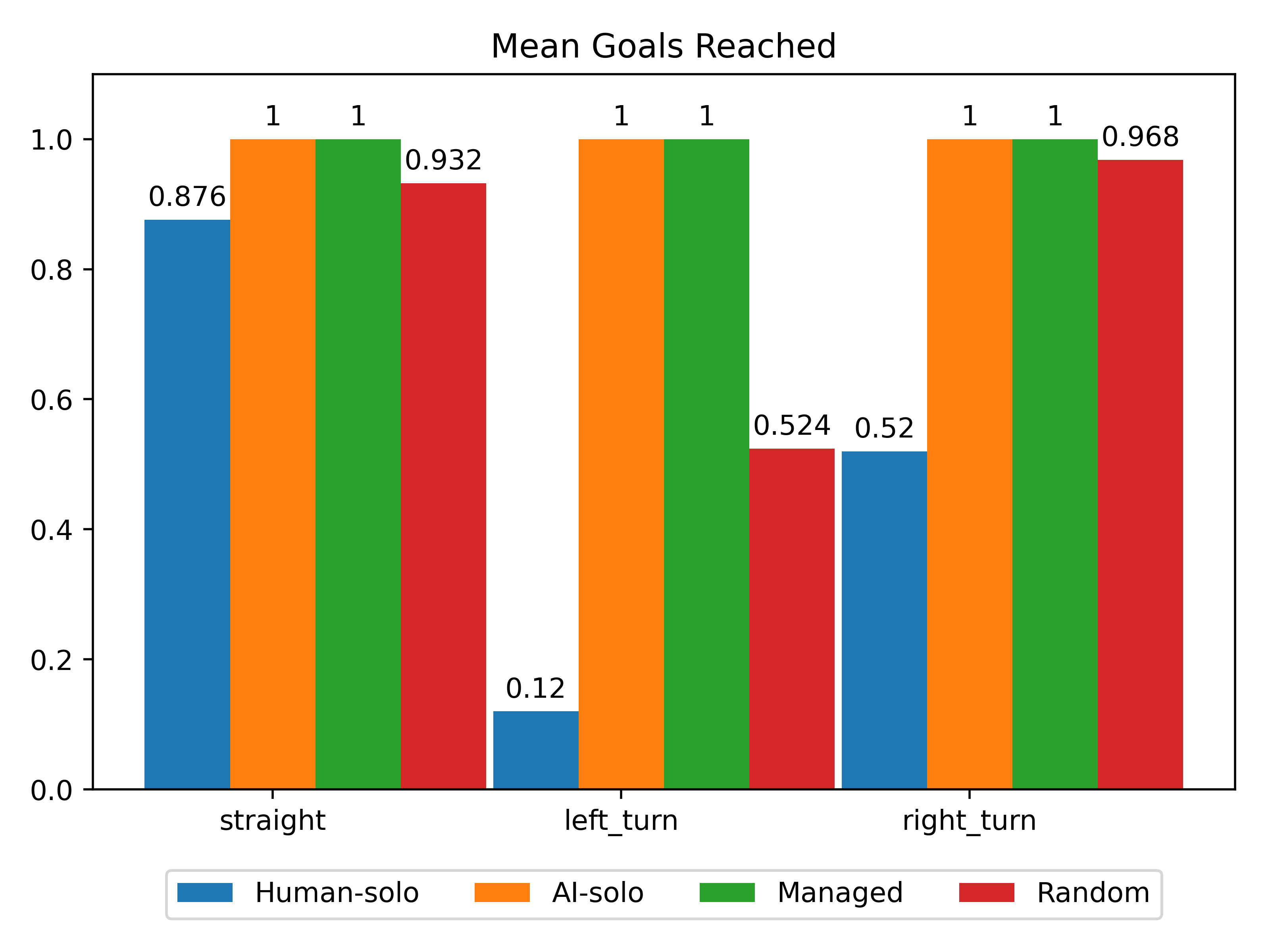}\vspace{-2mm}
        \caption{Night Driving and Error-free}
        \label{fig:h_light_ai_none1}
    \end{subfigure}
    \begin{subfigure}[t]{0.475\textwidth}
        \centering
        \includegraphics[width=0.875\textwidth]{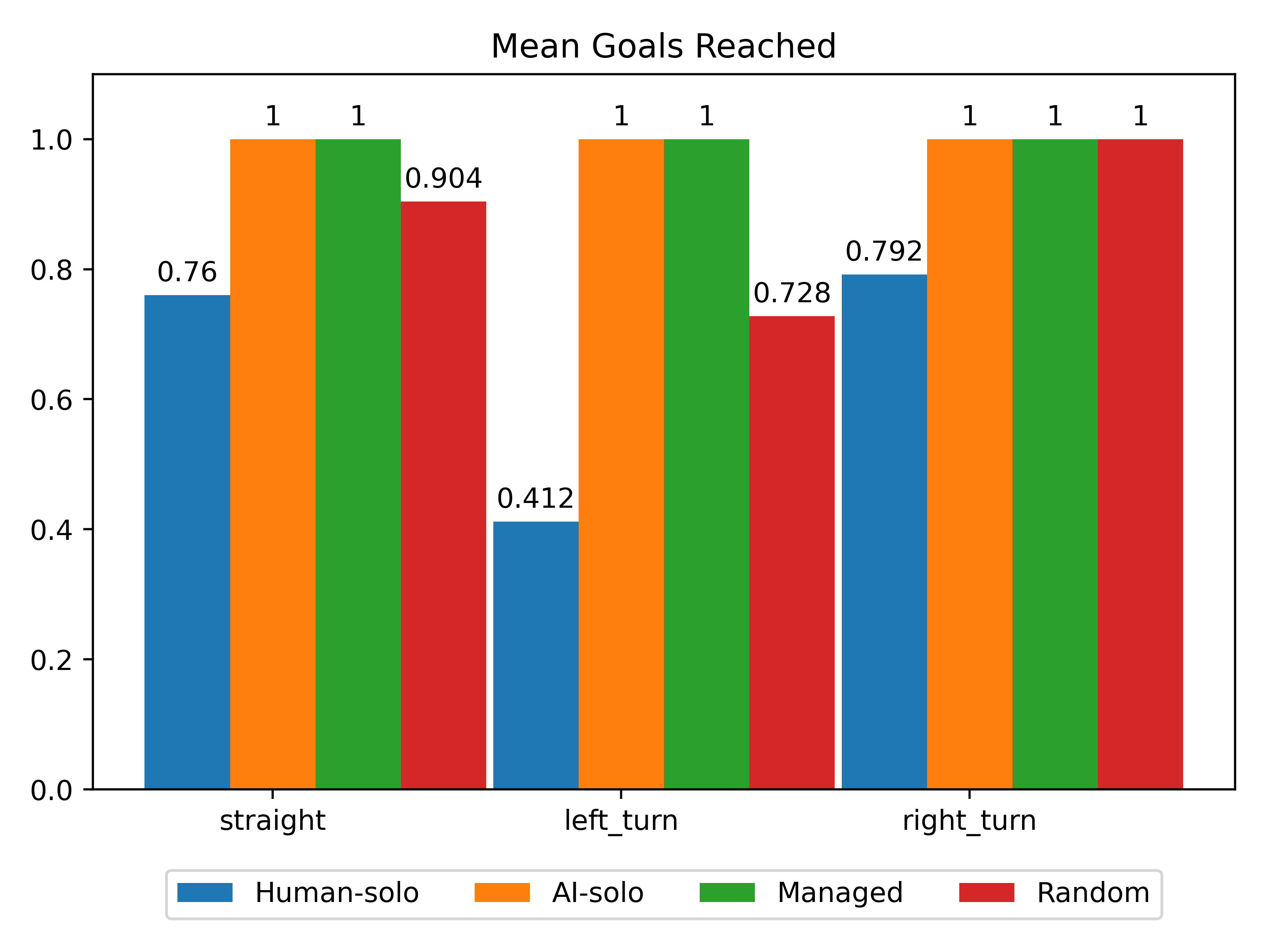}\vspace{-2mm}
        \caption{Night Driving and Error-free}
        \label{fig:h_light_ai_none2}
    \end{subfigure}\\\vspace{-0.25mm}
    \begin{subfigure}[t]{0.475\textwidth}
        \centering
        \includegraphics[width=0.875\textwidth]{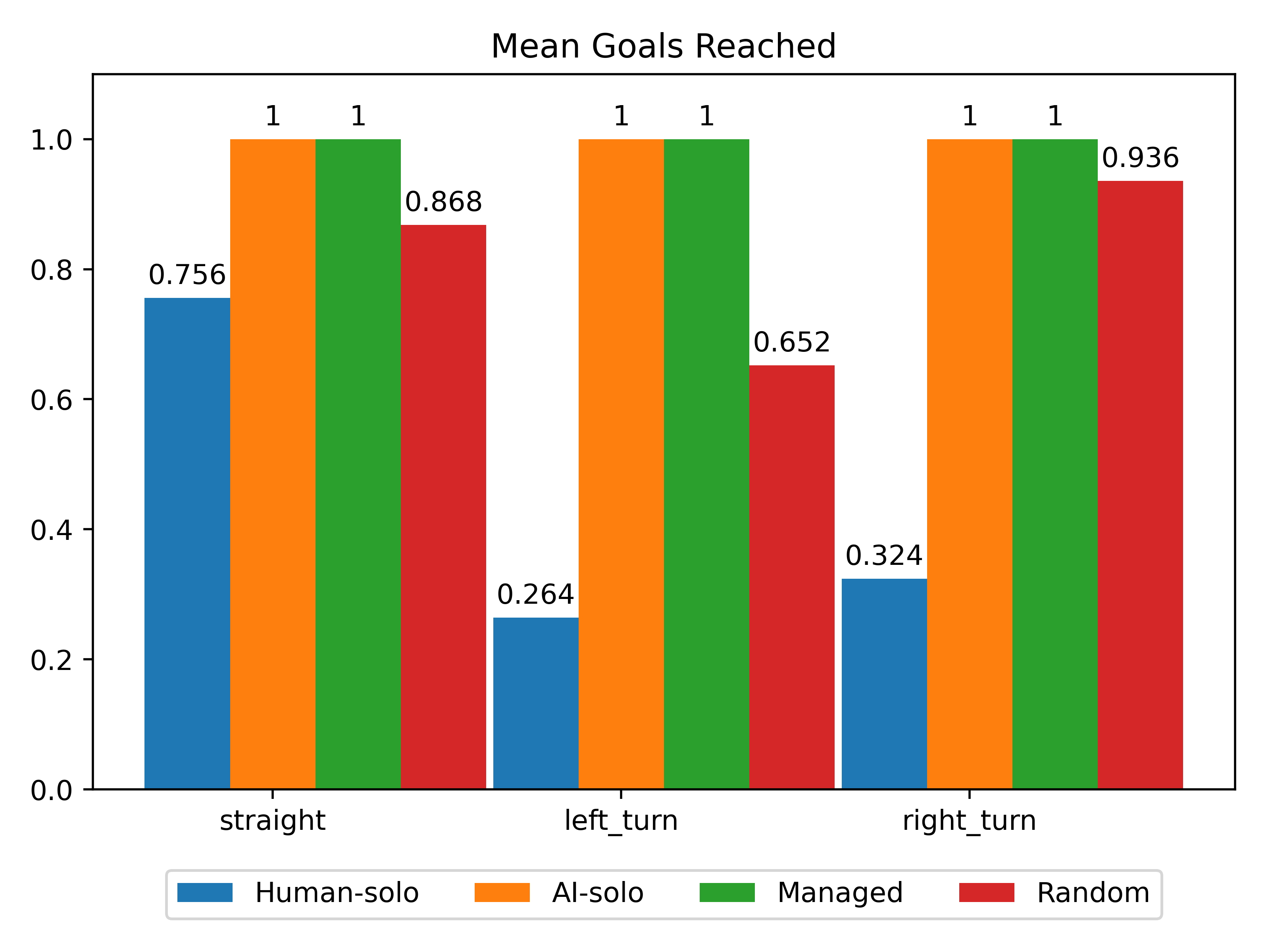}\vspace{-2mm}
        \caption{Distance-based Error and Error-free}
        \label{fig:h_weather_ai_none1}
    \end{subfigure}
    \begin{subfigure}[t]{0.475\textwidth}
        \centering
        \includegraphics[width=0.875\textwidth]{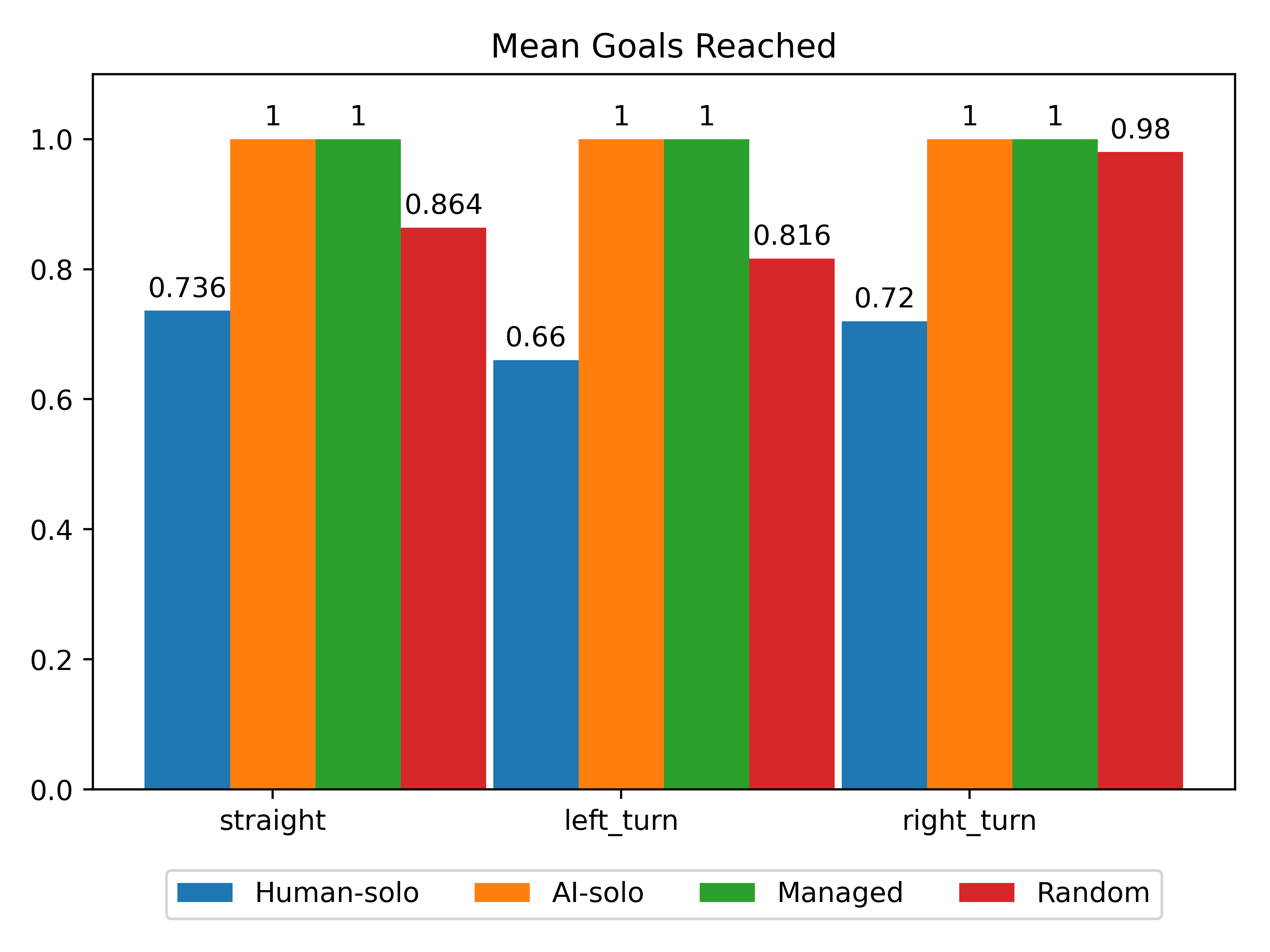}\vspace{-2mm}
        \caption{Distance-based Error and Error-free}
        \label{fig:h_weather_ai_none2}
    \end{subfigure}\\\vspace{-0.25mm}
    \begin{subfigure}[t]{0.475\textwidth}
        \centering
        \includegraphics[width=0.875\textwidth]{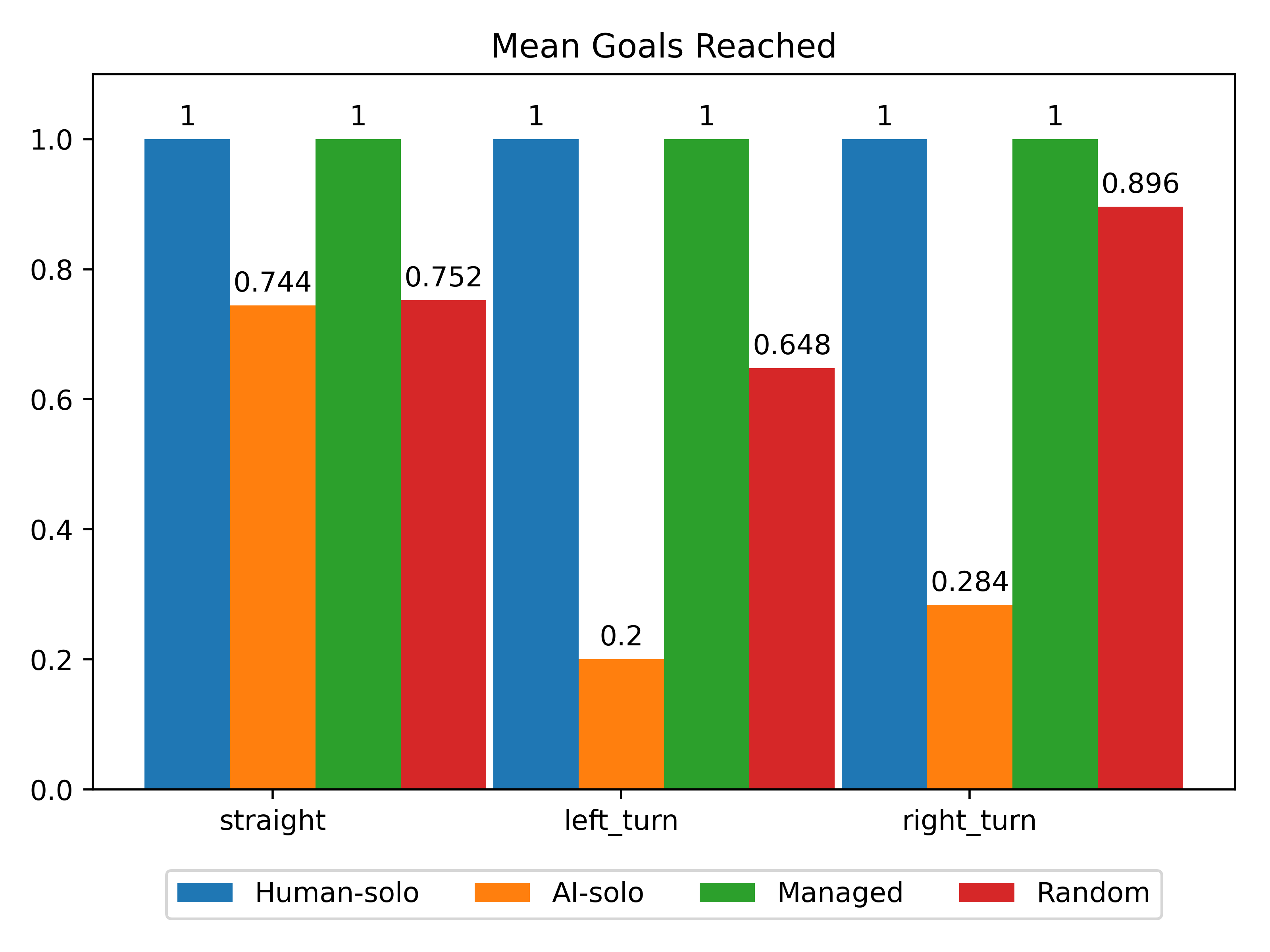}\vspace{-2mm}
        \caption{Color-based Error and Error-free}
        \label{fig:h_none_ai_color1}
    \end{subfigure}
    \begin{subfigure}[t]{0.475\textwidth}
        \centering
        \includegraphics[width=0.875\textwidth]{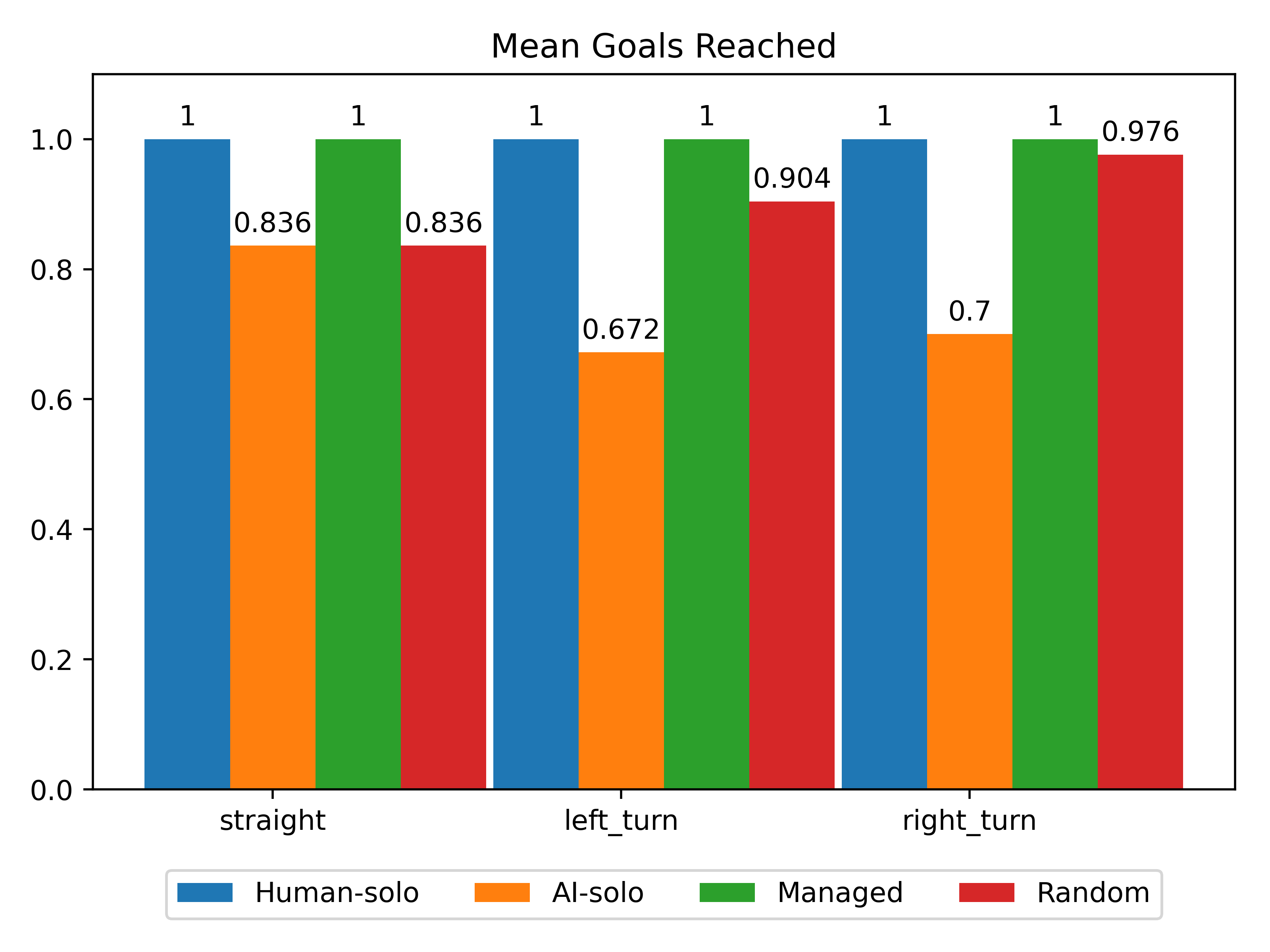}\vspace{-2mm}
        \caption{Color-based Error and Error-free}
        \label{fig:h_none_ai_color2}
    \end{subfigure}
    \vspace{-1.5mm}
    \caption{Performance comparison indicating success rates $n_g$ (i.e., goals reached) in each of the control cases -- human solo, AI solo, AI manager team, and random manager team. We also include mean intervention occurrences $\bar{n}_\beta$ in sub-figure (b). For figures (c) to (h), left-hand side plots show higher severity, while right-hand plots show lower severity for the same configuration of experiments.}
    \label{fig:biased_teams}
\end{figure}

For our training and tests, the two driving agents are configured with the same sensing parameters (e.g., observable distance), but we alter the perception of the human driver by masking all observations and preventing vehicle detection. As indicated by the results in Figure~\ref{fig:distracted_driver}, the manager successfully learns the predicted bias to the autonomous driver, resulting in $n_g$ scores matching the AI solo driver case. As an additional indication of manager behavior, the frequency of $\beta$ cues matches closely with the AI solo case. This indicates a close alignment of the AI solo and managed cases, which in turn indicates the manager having a strong bias toward the necessary agent. Combined, the results indicate successful manager training with the given team. Note, we will omit the delegation measurements $\bar{n}_\beta$ for the remaining cases as they are similar to the ones demonstrated in Figure~\ref{fig:distracted_driver_delegations}.

\subsubsection{Night Driving and Error-free}

For our next context, we demonstrate a scenario with one error-free driver and another which is hindered by the light-based context (e.g., driving at night). We demonstrate the manager performance with two mixtures of severity levels for the error-induced driver. The variation in severity is based on an increase/decrease in the context parameters. For night driving, we alter the severity levels by changing the darkness level, how quickly the perception degrades according to darkness as a function of distance, and the dimensions of the headlight region. For our experiments, we only alter the darkness level and degradation rate to ensure the headlight region has the same impact across experiments, so darkness level is the key factor. So, the increase of failure likelihood will be a direct result of the darkness outside the headlight region degrading perception. Specifically, the key difference in these experiments is how quickly the increase in brightness decays as a function of distance from the vehicle. Our experiments include an increase of $5\%$ in the rate of light degradation between the two severity levels.

In Figure~\ref{fig:h_light_ai_none1}, we see more severe variance in the impact the context has on the human driver. For instance, the performance in the straight and left turn cases goes from nearly error-free to nearly entirely errors. From our results, we see the manager learning a clear bias toward the error-free driver in order to ensure team success. As a reminder, we utilize a random action manager to compare performance when a delegation is made randomly. These delegations are again based on the outcomes of the current episode rather than from a prescribed path, so the performance of the team will depend on the driving agents and the current manager. In this scenario, we can see that the left turn and straight driving cases are of low consequence based on the random manager's performance. Clearly, the error-induced driver has specific intervals which are catastrophic but are also possible to avoid more easily. On the other hand, the left turn is a much more challenging task when it comes to ensuring proper delegation at a specific point in time. Therefore, we see a larger discrepancy in the managers' performances. The second context for this configuration sees a reduction in the severity of errors for most of the cases and a more consistent level of performance across the three driving tasks. From Figure~\ref{fig:h_light_ai_none2}, we can see again that our manager learns to rely on the error-free agent to achieve the best performance. Similar to the previous case, the straight driving and right turn cases offer more chances of team success than that of the left turn case. We again see our manager both identifying the desirable driver while also showing a larger margin from the performance of the random manager.

As a final note, there is some disparity between the solo human results which can appear to conflict with the context parameter severity levels. Following our discussion of Equation~\ref{eqn:path_failure}, we can attribute this to aspects of the driving task. The delegated agent makes its decision based on the vehicles it detects and whether it labels that vehicle as interacting. Therefore, the vehicle the agent bases its decisions on may change between episodes, or even time steps. In the case of night-driving, the team will have a headlight region where any vehicle in this region is detected with $100\%$ accuracy. If the vehicle the agent deems interacting lies outside the headlight region, the success of detection and corresponding decision will depend on the severity of the context. In the left-turn task, the time at which either vehicle is in the headlight region will depend on the previous actions of the delegated agent(s). Further, a missed detection of the ``actual'' interacting vehicle (i.e., the one leading to a failure/collision) will result in behavior which alters the critical points of an episode. Therefore, an agent which is potentially making an error with high confidence could lead to a case where there is no preventing a subsequent failure. From the results in Figure~\ref{fig:h_light_ai_none1} and Figure~\ref{fig:h_light_ai_none2}, we can see such an outcome most likely occurred. Despite the higher severity of the context parameters, the observation and detection was such that the detected vehicle reduced the likelihood of a failure. In the reduced severity case, the reduction likely resulted in a misguided importance given to the wrong vehicle, which had higher likelihood of detection due to the reduced severity. This then alters the team behavior and impacts the resulting team performance.

\subsubsection{Distance-based Error and Error-free}

Similar to the previous scenario, we again hinder one driver's performance via a context. Further, we similarly shift the severity to generate varying levels of difficulty. The severity increases by approximately $7\%$ regarding the degradation rate resulting from the distance to the potentially observed vehicle. In this scenario, the error-induced driver fails based on our distance-based degradation of sensing ability. Keeping with the previous scenario, we see the severity of difficulty being reduced for the straight and right turn cases. Regarding manager performance, we continue to see a strong indication the manager can recognize appropriate delegations in a biased team case. The results shown in Figure~\ref{fig:h_weather_ai_none1} correspond to a case of higher error severity. Despite this heightened error severity, the manager can recognize stronger choices. Keeping with our pattern of varying team performance and error severity, we demonstrate another case with a reduced severity of the errors for the context-dependent driver. Again, the results in Figure~\ref{fig:h_weather_ai_none2} indicate the trained manager recognizing the error-free agent's performance while the random manager performs around the mean of the two agents in most cases.

\subsubsection{Color-based Error and Error-free}

In our final biased team, we demonstrate a scenario with an error-free driver and another which fails to detect vehicles with colors similar to a particular ``blindness'' color. The severity of the error is based on how closely the blindness color aligns with the vehicle's color, so this error case will not depend on distance like the previous cases. In the case of color impacting perception, the severity shifts to increase the detection failure likelihood. For these experiments, we use parameters which should result in approximately $65\%$ and $75\%$ success in detection when the color matches the blindness color. This is further augmented in cases where the color is similar to the blindness color, which will slightly shift the impact on success according to the color distance outlined in Section~\ref{sec:masked_sensing}. The results in Figure~\ref{fig:h_none_ai_color1} represent the case of a more severe error level. The trained manager results again indicate the manager's ability to recognize the need to bias toward the error-free agent when making a delegation. For the results in Figure~\ref{fig:h_none_ai_color2}, we again reduce the severity of the error case. As expected, this makes little impact on the learned manager's performance but does reduce the likelihood of failure for the random manager. This is confirmed by the higher overall performance seen in the figure.

\subsection{Performance with both agents making mistakes}

In the following scenarios, we demonstrate teams with less apparent biases in driver performance. For these scenarios, each driver has their perception hindered by a context factor (e.g., night driving, weather/distance, etc.) to generate a team with a more varied concept of failure and success ability. The goal in this case is to provide a team for the manager which has no clear ``best'' agent in all cases. As such, the manager should learn an ideal mixture of the agents to best overcome their given perception deficiencies.

\subsubsection{Night Driving and Color-based Error}

\begin{figure}[htb]
    \centering
    \begin{subfigure}[t]{0.475\textwidth}
    \centering
        \includegraphics[width=0.875\textwidth]{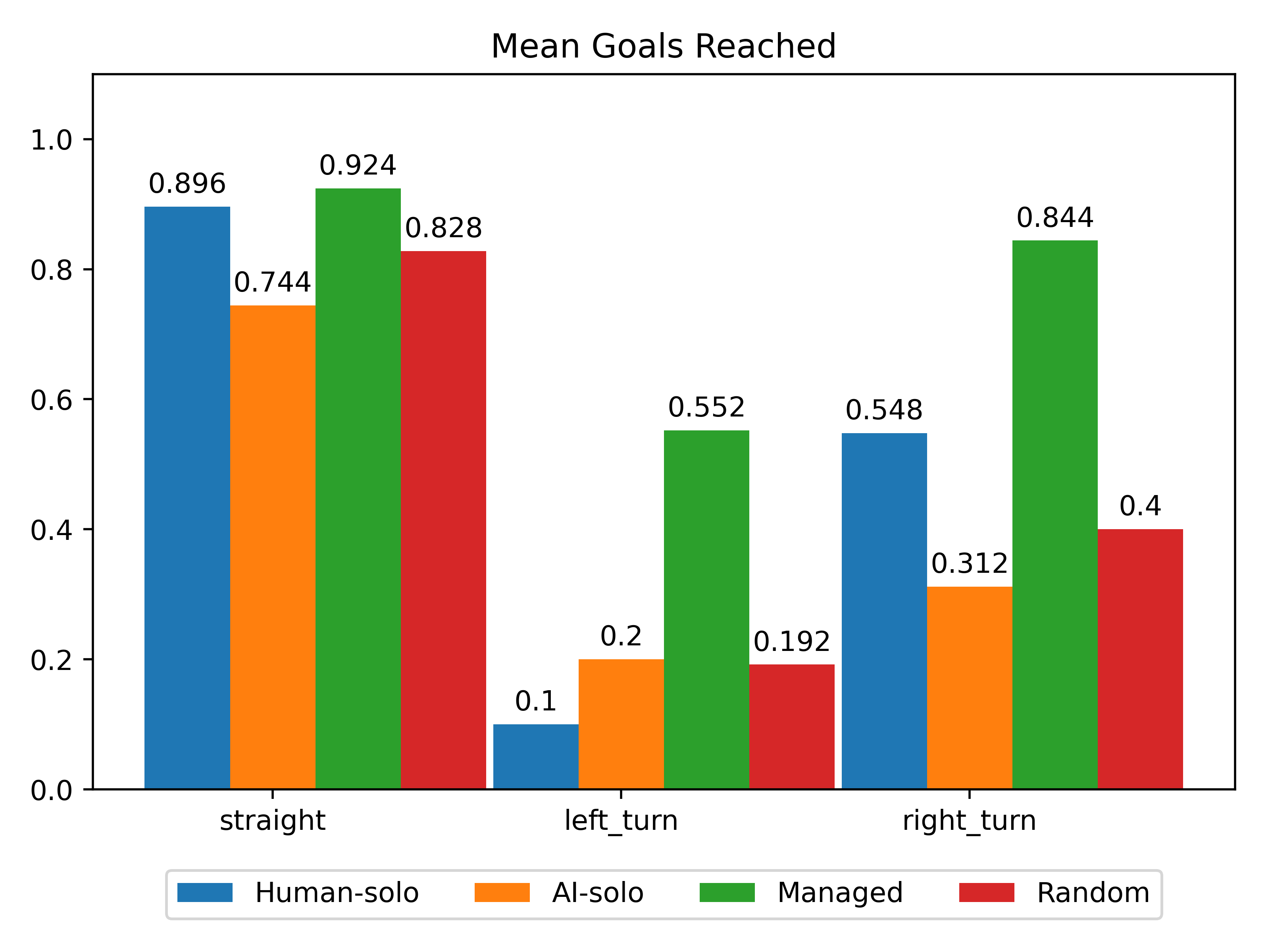}
        \caption{Night Driving and Color-based Error}
        \label{fig:h_light_ai_color1}
    \end{subfigure}
    \begin{subfigure}[t]{0.475\textwidth}
    \centering
        \includegraphics[width=0.875\textwidth]{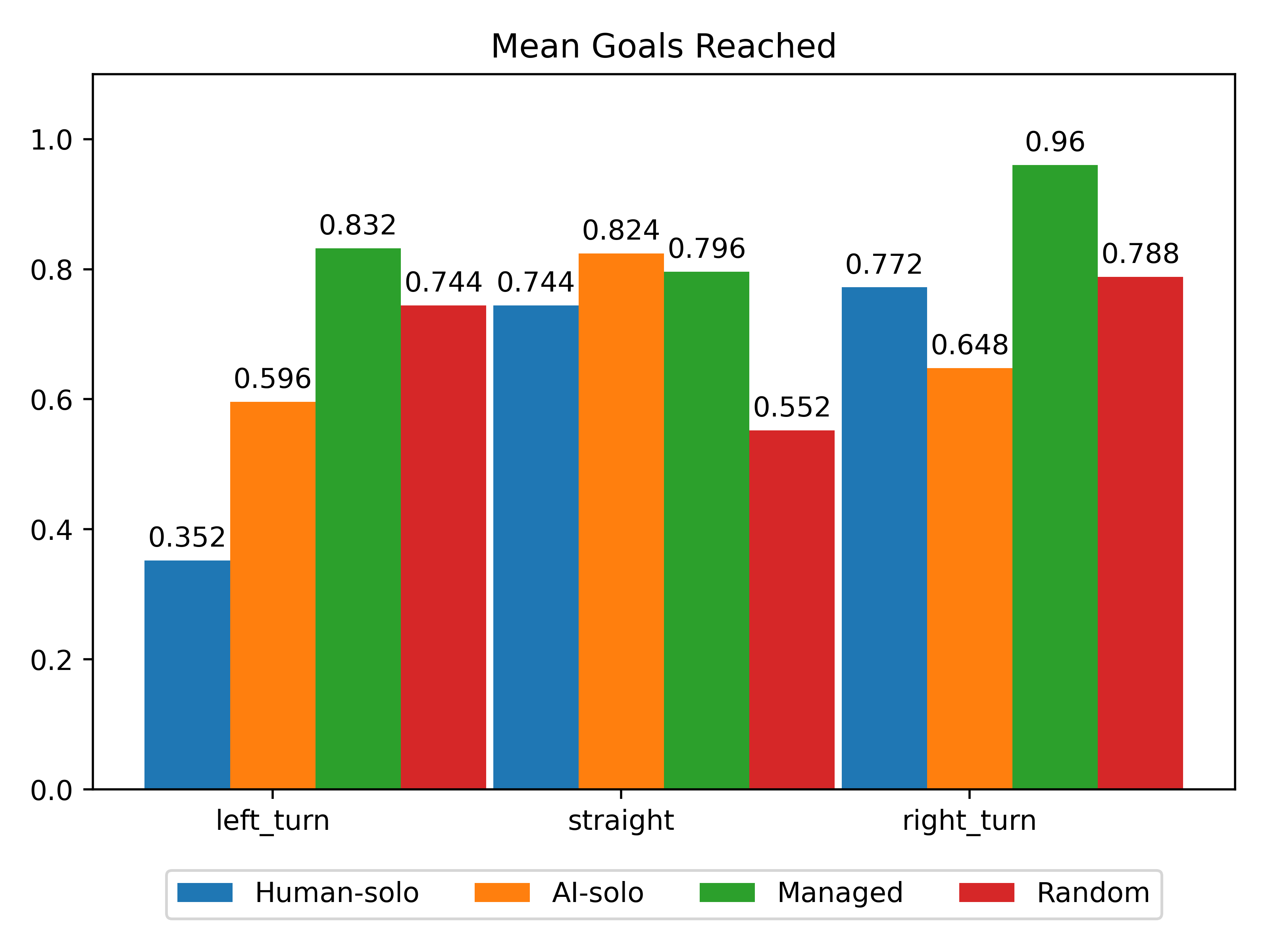}
        \caption{Night Driving and Color-based Error}
        \label{fig:h_light_ai_color2}
    \end{subfigure}\\
    \begin{subfigure}[t]{0.475\textwidth}
    \centering
        \includegraphics[width=0.875\textwidth]{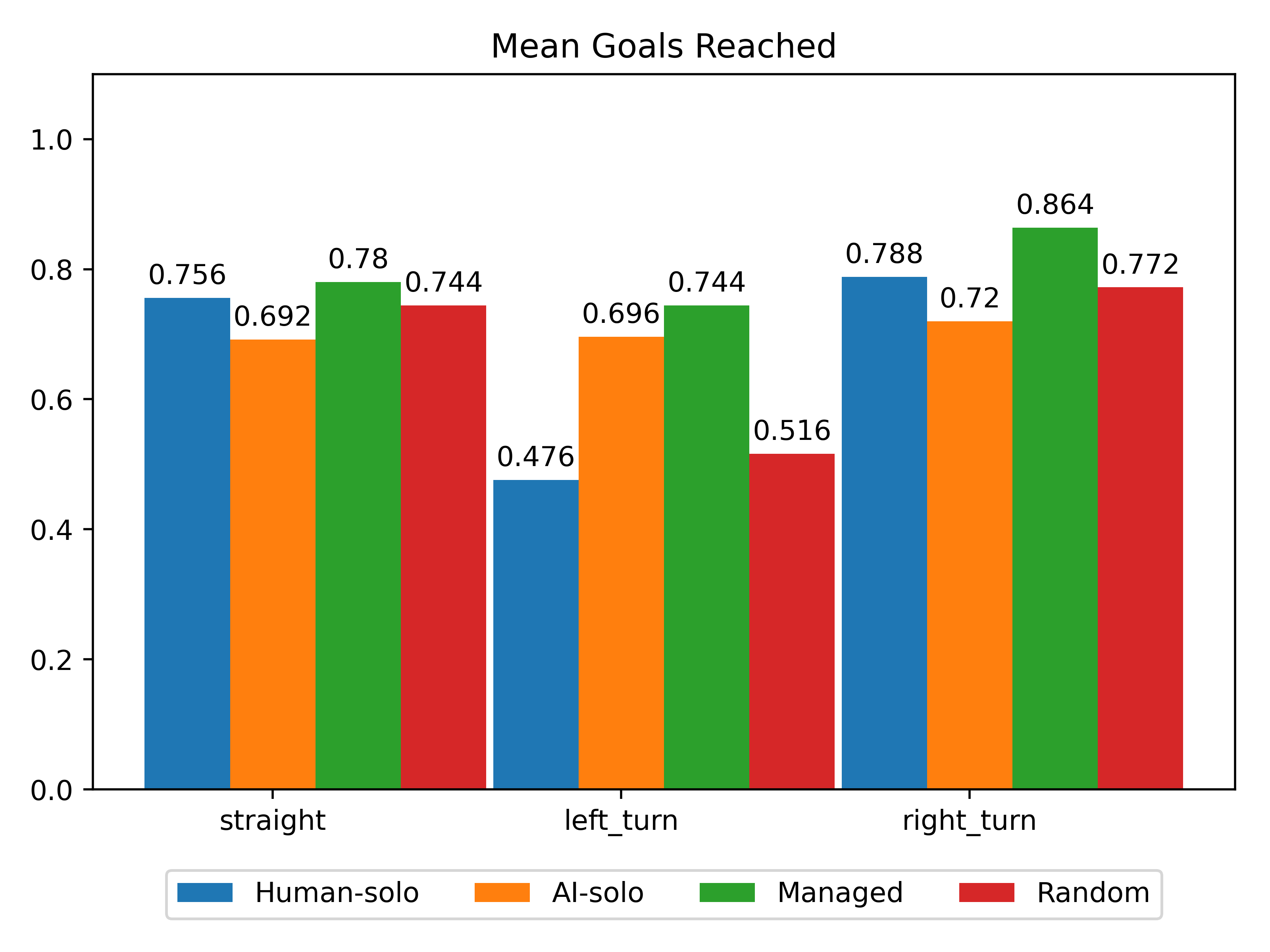}
        \caption{Night Driving and Distance-based Error}
        \label{fig:h_light_ai_fog1}
    \end{subfigure}
    \caption{Performance comparison indicating success rates $n_g$ in each of the control cases -- human solo, AI solo, AI manager team, and random manager team. Success is indicated by reaching the goal state within the episode time limit without collision with another vehicle.}
\end{figure}

In this configuration, we see a human driver which has its perception hindered by the low light of night driving. In this case, the perception decays based on the distance from the vehicle. The exception is the case of the area in front of the vehicle that is covered by the headlights. In this case, the perception is treated as the error-free context as there is no decay on perception in this region. For the AI driver, the perception is susceptible to failed detection when a vehicle has a color similar to its error color. Combined, these drivers have a mixture of failure types which are critical at different stages of the navigation of the environment. In cases where a car with a similar color to the error color is in the headlight region, the human driver will have no issue detecting the vehicle. Likewise, when any other vehicle is outside the headlight region, the AI driver will perform better in detection than the human driver. In Figure~\ref{fig:h_light_ai_color1} and Figure~\ref{fig:h_light_ai_color2}, we see the team performance in two mixtures of severity. These mixtures are based on the same parameters in our previous experiments and so we can demonstrate the same errors and severity levels, but we now demonstrate them in a mixed error team configuration. As indicated in both sets of results, the managed team does as approximately as well, and often better, than the case when the drivers are operating individually. Further, we see numerous cases of significantly higher performance of the learned manager over the use of a random manager. Each of these aspects indicate strong manager performance.

It is worth noting the occurrence of the managed team significantly outperforming random and solo cases. For example, the right turn case in Figure~\ref{fig:h_light_ai_color1} indicates the manager is able to exceed simply identifying a good agent to succeed in the task. In this case, the manager instead has found a sequence of interventions and delegations which result in a profound increase in team performance. Comparing the manager performance to the best solo agent, we see an increase of over $54\%$ for the right turn case and over $175\%$ for the left turn case. Comparing to the random manager, we see an increase of over $110\%$ for the right turn case and $187\%$ for the left turn. This indicates the manager's ability to explore combinations which can result in successful task completion and better avoidance of critical states which can lead to team failure. Hence, the manager goes beyond simply recognizing that one agent's performance is higher than another. In these results, we see the manager can find delegation choices which significantly improve the team performance. This indicates the manager can explore mixing between agents to fit the best-fitting agent to a particular part of the task. This fitting of agents is then completed for an entire trajectory which results in a significant increase in $n_g$ for the managed team. Relating to a context case, this could mean the manager found a delegation pattern which avoids issues for the night-based errors of the human driver by using the AI driver and then switching to the human driver when there is a vehicle that has a color close to the blindness color.

As an additional sample of performance, we generate a team with the human again subject to night driving and the AI driver having perception degrade with distance. In this case, the AI driver will have a similar susceptibility to failure as the human. Alternatively, the AI driver will not have an area like the headlight region which counteracts the perception failure. Seen in Figure~\ref{fig:h_light_ai_fog1}, the human and AI drivers both demonstrate moderate levels of error according to their parameters and often have similar overall performance. As indicated by the results, the combination under the manager provides an improvement in overall performance which exceeds individual performance. Further, the random manager performance is again such that there is a clear indication of the benefit of both a manager and a manager which is trained based on observation of driver performance.

\section{Conclusions and Future Work}\label{sec:conclusion}

We proposed the use of an RL-based manager for oversight of hybrid teams of agents which collaborate to accomplish a goal. The agents are expected to operate with only a single agent serving as the representative for the team. Authority is delegated to the authorized agent by the manager when the team reaches a point where a change in authority is needed (e.g., unsafe conditions). For our manager, we assume the teams will include a mixture of both agent types and skill levels. Additionally, we assume the various agents will have some potential for erroneous/undesirable behavior. The inclusion of potential agent failures ensures our manager is designed for a more realistic team. To indicate that a change in authority is needed, we assume constraints are provided which indicate acceptable limits on behavior. Failure to remain within these constraints will indicate the manager should make a new observation and make its next delegation decision. Our use of constraints allows us to link aspects of team performance to intervention and delegation decisions without requiring access to private information of any of the team's agents. The manager is designed to observe the performance of the agents, ensure the team maintains acceptable behavior, and to ensure the frequency of delegation changes is minimized. Further, we define a manager reward model which can link these aspects of performance in such a way as to also reduce the frequency of manager interventions while still promoting team success. Reduced intervention is accomplished and refined by not only penalizing for interventions but by scaling the penalties according to their importance indicated by the constraints.

Our framework also offers a method of managing multiple agents over extended time windows without direct observation of behavior for the manager. As our manager is only providing oversight, the manager ignores all states where the intervention constraints are not violated. Therefore, our manager can remain in the background for the team whenever possible. Further, the manager's exploration space is reduced to only those states in which an intervention is required. This helps reduce the exploration task for the manager and simplifies the search for optimal behavior. Further, this ensures the manager is compatible with agents which have already learned a model of behavior prior to operating with the team. This is important as many models of behavior are trained individually, which is especially true for humans as they would most commonly learn to accomplish most of their daily tasks without any expectation of operation with an artificial agent. Therefore, our framework supports teams of individual and unique agents which have no expectations regarding each other. Under this framework, we can generate a Reinforcement Learning (RL) model which conforms to traditional RL theory. As such, we show that our model converges to the optimal solution in the discrete case.

We demonstrate the operations of the general framework in a concrete relevant case. Manager utility and performance is highlighted in a simulated driving task with a mixture of human and AI drivers of varying skill/reliability levels. The manager performance is measured from the perspective of driver selection under constraints indicating whether a vehicle has entered an unsafe state. Given the performance indicated by our results, we see the manager clearly learning a relationship between the delegation choices, intervention cases, and overall performance. In all contexts, we see the manager learning delegation patterns which result in performance at least as good as the best performing agent and, most typically, significantly outperforming the best solo agent. As many of the contexts resulted in agents with significant chances of failure, we expected a reduction in failures without anticipating the complete removal of errors. Our results clearly demonstrate such a pattern, indicating a manager which can learn to better delegate authority while operating as an outside observer. Therefore, our model can use a set of constraints to dictate desirable team operation while reducing assumptions regarding the private information available to the manager during the decision-making process.

For a general assessment of our model and results, we indicate the following. First, the use of constraints to isolate manager decisions from agent information means the manager is not given a method by which it can measure the skill of a team member. The manger can only assess through observation whether an agent is making decisions which conform to its constraints. Second, our experimental scenario(s) are based on a 2D simulation of driving, which reduces the representational complexity of the environment. Training and testing in a more realistic setting (e.g., robotics, 3D driving simulator, etc.) could better illustrate the suitability of our model in real-world settings. As for positive aspects, we demonstrate our model serves to learn effective management of erroneous agents in a domain with continuous states and agent actions. As such, we demonstrate our manager can learn how to manage teams in a complex domain with loose associations between single decisions and overall outcomes. Further, our introduction of manager constraints has shown effective in indicating necessary manager interventions while eliminating the need for direct observations of team decisions and knowledge. Therefore, our manager demonstrates effective decisions with significantly reduced access to team-related information. As such, we present a highly effective manager with realistic assumptions for manager observations.

Extending from our assessment, we consider several key topics for future work. First, our model should be tested in more complex and realistic domains such as 3D driving simulators (e.g., CARLA \cite{Dosovitskiy17}). These would illustrate whether our model scales to domains with higher complexity in representation and dynamics. Additionally, this would further indicate the suitability of our model in real domains. As an additional extension, experiments with real human users would allow us to investigate the suitability and acceptance of our model with the target users. We could assess our model's effectiveness and the perceived performance by the human users. This could indicate any additional factors which need to be modified to increase the compatibility with an RL manager with human team members. Finally, we could extend our approach to cases with either multiple agents operating concurrently, or we could consider cases of multiple teams operating in the same domain. In such cases, our model would need to be extended to account for multiple actions and disjoint agent states. In other words, agents operating independently would have their own contribution to changes in their, or the team's, state in the environment. Such extensions would require a broader awareness of multiple agent states and decisions. Alternatively, unobserved aspects of the dynamics could lead to requiring representation using modeling techniques such as Partially Observable Markov Decision Processes (POMDP) \cite{sutton2018reinforcement}.

\begin{acks}
This work was partially supported by the \grantsponsor{CHIST-ERA}{CHIST-ERA-19-XAI010 SAI}{https://www.chistera.eu/projects/sai} 
project. M. Conti’s and A. Passarella's work was partly funded by the \grantsponsor{PE0000013}{PNRR - M4C2 - Investimento 1.3, Partenariato Esteso PE00000013 - "FAIR"}{https://future-ai-research.it/} funded by the European Commission under the NextGeneration EU programme.
\end{acks}

\printbibliography

\newpage

\appendix

\section{Markov Decision Process (MDP)}\label{sec:mdp}

In Reinforcement Learning (RL), we assume the agent operates with respect to a Markov Decision Process (MDP). Therefore, we assume a standard definition of the MDP $M = \langle S, A, R, T, \gamma\rangle$, where:
\begin{itemize}
    \item $S$, the state space, is the set of states the agent can traverse
    \item $A$, the action space, is the set of actions an agent is allowed to take
    \item $R: S\times A\rightarrow\mathbb{R}$, is the reward function denoting action utility
    \item $T: S\times A\times S\rightarrow[0, 1]$, the transition function, denotes the probability of transitioning between two states when an action is performed
    \item $\gamma$, the discount parameter, dictates how strongly values decay according to their temporal distance
\end{itemize}
These components of the MDP model define the states $s\in S$ of the environment the agent can reach through sequences of actions $a\in A$. These actions will result in transitions between states with probability according to the transition function $T$. The dynamics specified by $S, A,$ and $T$ allow the agent to interact with the environment while $R$ indicates how desirable the behaviors are for the given MDP. When learning a behavior policy, the values specified by $R$ teach an agent to accomplish the underlying task in such a way as to best fit the values imposed by $R$. Commonly, an agent will learn behaviors which will maximize the likely utility of their sequence of choices. Finally, the learning process is impacted by the discount parameter $\gamma\in[0, 1]$ which indicates how strongly values of states reachable from the current state impact the estimated value of the current state and actions.

A state's value is based on both the immediate reward possible by actions taken in that state and the value of states reachable from it. Therefore, the discount $\gamma$ will indicate the severity with which these future state values are reduced. A higher value of $\gamma$ will increase the strength of the additional value of states which are reachable through a chain of transitions from the current state. These discounted estimates of value will indicate utility to the agent as it learns to optimize its behavior accordingly.

\section{Proof of Convergence}\label{appendix:q_convergence}

Following the proof of convergence for Q-Learning on MDPs in \cite{jaakkola1993convergence,melo2001convergence}, we will show that the learning algorithm converges to the optimal state-action value function:
\begin{equation}
    Q^*(s_j,d) = \sum_{s_k\in S_\BR}b^{\pi_d}_{jk}\left[R_M + \gamma V^*(s_k)\right].
\label{eqn:optimal_Q}
\end{equation}
\begin{proof}
    As with traditional Q-Learning, we will start by showing that the optimal $Q$ is a fixed point of a contraction operator $\Hq$. First, recall that we are considering an Intervening MDP $M_I = \langle S, S_\BR, A_M, R_M, T_M, D, \beta, \gamma\rangle$. Additionally, recall that under the AMC-based model of transitions between intervention states, we model the probabilities of transitioning from one intervening state to another as 
    \begin{equation}
        P(s_j|s_i,d) = b^{\pi_d}_{ij} = \left[\left(I - \BQ\right)^{-1}\BR\right]_{ij}
    \end{equation}
    Starting with the single step model
    \begin{equation}
        P = \begin{bmatrix}
            \begin{array}{c:c}
                \BI_{k-j} & \BO_{(k-j)\times j} \\
                \hdashline
                \BR_{j\times(k-j)} & \BQ_{j\times j}
            \end{array}
        \end{bmatrix}
    \end{equation}
    we can extend this to the models of $n$-step transitions (i.e., $P^n$). This leads to the higher powers of $\BQ$ which indicate the number of times a transient state is visited prior to reaching an absorbing state. This representation is then extended to account for all possible step numbers via 
    \begin{equation}
        \sum_{k=0}^\infty\BQ^k = \left(I - \BQ\right)^{-1}
    \end{equation}
    With this model, we therefore see our use of $b^{\pi_d}_{jk}$ provides a valid state transition probability given infinite horizon in a finite AMC.
    
    Now, define
    \begin{equation}
        (\Hq q)(s_j, d) = \sum_{s_k\in S_\BR}b^{\pi_d}_{jk}\left[R_M(s_j,d,s_k) + \gamma\max_{d'}q(s_k,d') \right]
    \end{equation}
    We show $\Hq$ is a contraction in the sup-norm by showing that $\norm{\Hq q_1 - \Hq q_2}_\infty \leq \gamma\norm{q_1 - q_2}_\infty$. From the definition of $\Hq$, we see that
    \begin{align*}
        \norm{\Hq q_1 - \Hq q_2}_\infty &= \max_{s_j, d}\left|\sum_{s_k\in S_\BR}b^{\pi_d}_{jk}\left[R_M(s_j,d,s_k) + \gamma\max_{d'}q_1(s_k,d') -\, R_M(s_j,d,s_k) - \gamma\max_{d'}q_2(s_k,d')\right]\right|\\
        &= \max_{s_j, d}\gamma\left|\sum_{s_k\in S_\BR}b^{\pi_d}_{jk}\left[\max_{d'}q_1(s',d') - \max_{d'}q_2(s',d')\right]\right|\\
        &\leq \max_{s_j, d}\gamma\sum_{s_k\in S_\BR}b^{\pi_d}_{jk}\left|\max_{d'}q_1(s',d') - \max_{d'}q_2(s',d')\right|\\
        &\leq \max_{s_j, d}\gamma\sum_{s_k\in S_\BR}b^{\pi_d}_{jk}\max_{d'}\left|q_1(s',d') - q_2(s',d')\right|\\
        &= \max_{s_j, d}\gamma\sum_{s_k\in S_\BR}b^{\pi_d}_{jk}\norm{q_1 - q_2}_\infty\\
        &\leq \gamma\norm{q_1 - q_2}_\infty
    \end{align*}
    Note the final inequality follows from the AMC model outlined above. With the Definition of $\BQ$ under the AMC model, the probabilities of transitioning from a transient state to one of the absorbing states in n steps goes to one as $n\rightarrow\infty$, giving an upper bound on $b^{\pi_d}_{jk}$. Finally, this ensures the above inequality holds and that $\Hq$ satisfies the properties of a contraction. Given $\Hq$ is a contraction, we will then show that $Q_t$ converges to $Q^*$ as $t\rightarrow\infty$. To start, we will assume that for $\forall t$, we have $0\leq\alpha_t\leq 1, \sum_t\alpha_t=\infty, \mathrm{ and } \sum_t\alpha_t<\infty$. Let
    \begin{equation*}
        F_t(s,d) = R_M(s,d,X(s,d)) + \gamma\max_{d'}Q_t(X(s,d),d') - Q^*(s,d)
    \end{equation*}
    for next state distribution $X(s,d)$. Then, we will first show that $\norm{\eval{F_t(s,d)|\mathcal{F}_t}}_\infty \leq\gamma\norm{\Delta_t}_\infty$, with $\gamma < 1$. Let
    \begin{equation*}
        \Delta_t = Q_t(s,d) - Q^*(s,d)
    \end{equation*}
    then
    \begin{equation*}
        \Delta_{t+1} = (1 - \alpha_t)\Delta_t + \alpha_tF_t(s,d)
    \end{equation*}
    Therefore,
    \begin{align*}
        \eval{F_t(s_j,d)|\mathcal{F}_t} &= \sum_{s_k\in S_\BR}b^{\pi_d}_{jk}\left[R_M(s_j,d,s_k) + \gamma\max_{d'}Q_t(s_k,d') - Q*(s_j,d)\right]\\
        &= (\Hq Q_t)(s_j,d) - Q^*(s_j,d)\\
        &= (\Hq Q_t)(s_j,d) - (\Hq Q^*)(s_j,d)\\
        &\overset{(a)}{\Rightarrow} \norm{\eval{F_t(s_j,d)|\mathcal{F}_t}}_\infty \leq\gamma\norm{Q_t - Q^*}_\infty\\
        &= \gamma\norm{\Delta_t}_\infty
    \end{align*}
    with $(a)$ resulting from the contraction proof. Next, we will show that $\V{F_t(s)|\mathcal{F}_t} \leq C(1 + \norm{\Delta_t}_\infty)^2$ for $C > 0$.
    \begin{align*}
        \V{F_t(s)|\mathcal{F}_t} &= \eval{\left(R_M(s,d,X(s,d)) + \gamma\max_{d'}Q_t(s',d') - Q^*(s,d) - (\Hq Q_t)(s,d) + Q^*(s,d)\right)^2}\\
        &= \eval{\left(R_M(s,d,X(s,d)) + \gamma\max_{d'}Q_t(s',d') - (\Hq Q_t)(s,d)\right)^2}\\
        &\overset{(a)}{=} \V{R_M(s,d,X(s,d)) + \gamma\max_{d'}Q_t(s',d')|\mathcal{F}_t}\\
        &\leq C(1 + \norm{\Delta_t}_\infty)^2
    \end{align*}
    with the final inequality resulting from \cite{melo2001convergence}. Further, note that $(a)$ comes from the fact that $(\Hq Q_t)(s,d) = \eval{Q_t(s,d)}$. Therefore, we see that $Q^t$ converges to $Q^*$ with probability $1$.
\end{proof}

\end{document}